\documentclass[letterpaper]{article} 
\usepackage{aaai23}  
\usepackage{times}  
\usepackage{helvet}  
\usepackage{courier}  
\usepackage[hyphens]{url}  
\usepackage{graphicx} 
\usepackage{natbib}  
\usepackage{caption} 
\usepackage{algorithm}
\usepackage{algorithmic}
\usepackage{mathtools}

\usepackage{newfloat}
\usepackage{listings}
\DeclareCaptionStyle{ruled}{labelfont=normalfont,labelsep=colon,strut=off} 
\floatstyle{ruled}
\newfloat{listing}{tb}{lst}{}
\floatname{listing}{Listing}
\usepackage[utf8]{inputenc} 
\usepackage{hyperref}       
\usepackage{url}            
\usepackage{booktabs}       
\usepackage{amsfonts}       
\usepackage{nicefrac}       
\usepackage{microtype}      
\usepackage{soul}
\usepackage{arydshln}
\usepackage{lipsum,booktabs}
\usepackage{lipsum}
\usepackage{times}
\usepackage{vwcol}
\usepackage{booktabs,multirow}  
\usepackage{amsfonts}       
\usepackage{nicefrac}      
\usepackage{microtype}     
\usepackage{multirow}
\usepackage{algorithm}
\usepackage{mathtools}
\usepackage{multicol}
\usepackage[misc]{ifsym}
\usepackage{pifont}

\title{AdaTask: A Task-aware Adaptive Learning Rate Approach to Multi-task Learning}

\author {
    Enneng Yang \textsuperscript{\rm 1}\equalcontrib \thanks{This work was done during an internship at Tencent.},
    Junwei Pan \textsuperscript{\rm 2}\footnotemark[1], 
    Ximei Wang \textsuperscript{\rm 2}, 
    Haibin Yu \textsuperscript{\rm 2}, 
    Li Shen \textsuperscript{\rm 3}\\
    Xihua Chen \textsuperscript{\rm 2}, 
    Lei Xiao \textsuperscript{\rm 2}, 
    Jie Jiang \textsuperscript{\rm 2}, 
    Guibing Guo \textsuperscript{\rm 1}\thanks{Corresponding author.}
}
\affiliations {
    \textsuperscript{\rm 1} Northeastern University, China \\
    \textsuperscript{\rm 2} Tencent Inc, China \\
    \textsuperscript{\rm 3} JD Explore Academy, China\\
    {ennengyang@stumail.neu.edu.cn, mathshenli@gmail.com, guogb@swc.neu.edu.cn \{jonaspan,messixmwang,nathanhbyu,tinychen,shawnxiao,zeus\}@tencent.com}
}

\begin{document}
\maketitle

\begin{abstract}
Multi-task learning (MTL) models have demonstrated impressive results in computer vision, natural language processing, and recommender systems. Even though many approaches have been proposed, how well these approaches balance different tasks on each parameter still remains unclear. 
In this paper, we propose to measure the task dominance degree of a parameter by the total updates of each task on this parameter. Specifically, we compute the total updates by the \textit{exponentially decaying \textbf{A}verage of the squared \textbf{U}pdates} (AU) on a parameter from the corresponding task.
Based on this novel metric, we observe that many parameters in existing MTL methods, especially those in the higher shared layers, are still \textit{dominated} by one or several tasks. The dominance of AU is mainly due to the dominance of accumulative gradients from one or several tasks. Motivated by this, we propose a Task-wise Adaptive learning rate approach, AdaTask in short, to separate the \emph{accumulative gradients} and hence the learning rate of each task for each parameter in adaptive learning rate approaches (e.g., AdaGrad, RMSProp, and Adam). Comprehensive experiments on computer vision and recommender system MTL datasets demonstrate that AdaTask significantly improves the performance of dominated tasks, resulting SOTA average task-wise performance.  Analysis on both synthetic and real-world datasets shows AdaTask balance parameters in every shared layer well.
\end{abstract}

\section{Introduction} \label{intro}

Multi-task learning (MTL) has emerged as a promising approach to jointly learning multiple tasks together by sharing structure across tasks in computer vision~\cite{Cross-Stitch,mtl_cv_taskonomy,2019cvprdwa}, natural language processing~\cite{mtl_nlp_translation,mtl_nlp_lm}, and recommender systems~\cite{mmoe,MT-FwFM,ple}. However, due to optimizing different tasks simultaneously, MTL is prone to be dominated by one or several tasks, leading to the performance deterioration of other tasks~\cite{ple}. 

While there have been a significant number of approaches to MTL, it is still unclear \textit{to what extent these approaches resolve the task dominance of parameter optimization during MTL training}. This is due to the fact that there are currently no metrics to quantify the degree of task dominance in MTL. To fill this gap, in this paper, we propose to calculate the total updates of a parameter from a specific task by the exponentially decaying Average of the squared parameter Updates (AU in short) from this task. The dominance of a task on a given parameter is then measured by the ratio of AU from this task against the overall AU from all tasks. Intuitively, for a parameter, if a task has a much larger AU ratio than other tasks, then this parameter can be regarded as being dominated by this task. Based on this metric, we observe that all existing MTL methods still suffer from task dominance on shared parameters, especially on those in higher shared layers.

We suspect that the dominance of accumulative gradients may be one of the main reasons for the task dominance issue in MTL. Modern adaptive learning rate optimizers commonly used in deep learning adaptively adjust the learning rate of each parameter during the model training process, which is usually better than the gradient descent optimizer with a global learning rate (see Appendix F for details). 
The adaptive learning rate is usually proportional to the inverse of the accumulative gradients, and both the learning rate and accumulative gradients are computed across all tasks for shared parameters in MTL. Therefore, for a given parameter, if a task provides much larger gradients than other tasks, then this task would dominate both the overall accumulative gradients and the learning rate.

Motivated by this, we propose AdaTask to separate the accumulative gradients of each task for each shared parameter in MTL optimizers. By doing this, no task would dominate the overall accumulative gradients as well as the learning rate anymore since each task accumulates and computes its own ones. Almost all modern adaptive learning rate optimizers and their variants \cite{adagrad,ADADELTA,adam,amsgrad,adabelief,zou2019sufficient,chen2022efficient,chen2022towards,chen2021quantized,zou2018weighted}, such as AdaGrad, AdaDelta, RMSProp, Adam, AMSGrad, etc., can utilize AdaTask to resolve the task dominance issue in MTL training. The contribution of this work is summarized as follows:
\begin{itemize}
    \item[(1)] We quantify the task dominance of a parameter by the task-wise exponentially decaying average of the squared updates(AU) in MTL and observe that existing MTL methods still suffer from task dominance.
    \item[(2)] We propose a Task-aware Adaptive Learning Rate method (AdaTask) to separate the accumulative gradients of each task during optimization in the MTL models. 
    \item[(3)] Extensive experiments on the synthetic and three public datasets from the CV and recommendation demonstrate that AdaTask significantly improves the performance of the dominated task(s), while achieving SOTA average task-wise performance. Analysis shows that AdaTask balances parameters in all shared layers well.
\end{itemize}

\section{Related Work}
\label{sec:related_work}

\noindent
\textbf{Multi-Task Learning Architecture.}
Shared Bottom~\cite{shared_bottom} is the most basic and commonly used MTL structure. With hard parameters shared across tasks, it suffers from the negative transfer. Cross-stitch network~\cite{Cross-Stitch} and Sluice network~\cite{SluiceNetworks} were proposed to learn weights of linear combinations to fuse representations from different tasks selectively. However, the weights for different tasks are static for all samples. MOE~\cite{moe} first proposed to share some experts at the bottom and combine experts through a gating network. MMOE~\cite{mmoe} extends MOE to utilize different gates for each task to obtain different fusing weights in MTL. Similarly, MRAN~\cite{zhao2019multiple} applied multi-head self-attention to learn different representation subspaces for different feature sets. PLE~\cite{ple} develops both shared experts and task-specific experts, together with a progressive routing mechanism to further improve learning efficiency. In addition, there have been some studies of the adaptive selection of layers or neurons instead of experts.
AdaShare~\cite{adashare} selects some layers for each task from multiple shared layers.
TAAN~\cite{liu2020adaptive} adopts a task adaptive activation network to enable flexible and low-cost MTL.
MTAN~\cite{2019cvprdwa} uses the attention mechanism to extract the feature representation corresponding to the task from the shared backbone. In contrast to the above, which focuses on the shared module design, some works explicitly accomplish knowledge transfer between tasks at the output layer. ESSM~\cite{ESMM} models the sequential pattern impression of $\rightarrow$ click $\rightarrow$ conversion. CrossDistil~\cite{crossdistill2022} proposes to enhance knowledge transfer between tasks through distillation.

\noindent
\textbf{Multi-Task and Multi-Objective Optimization.}
There are also many MTL works to revolve negative transfer by either balancing the gradients of all tasks~\cite{uncertaintyweighting,2019cvprdwa,LBTWAAAI2019,chen2018gradnorm} or avoiding gradient conflicts~\cite{gradientsurgery,graddrop,cagrad-liu2021conflict}. UW~\cite{uncertaintyweighting} calculates the weights based on homoscedastic uncertainty, while DWA~\cite{2019cvprdwa} and LBTW~\cite{LBTWAAAI2019} are based on the rate of change of loss for each task. GradNorm~\cite{chen2018gradnorm} proposed adding a regularization term to the loss function to balance the training rates of all tasks. GradientSurgery~\cite{gradientsurgery} proposes gradient surgery to avoid interference between task gradients. GradDrop~\cite{graddrop} samples gradients based on their level of consistency. CAGrad~\cite{cagrad-liu2021conflict} seeks the gradient update direction by maximizing the task with the least loss reduction. In addition, some works formulate multi-task learning as a multi-objective optimization problem and seek discrete or continuous Pareto optimal solutions~\cite{mtlasmooSenerK18, ma2020efficientparetoICML2020, www2021_paretors}. Finally, some works in MTL attempt to balance the interference of auxiliary tasks with the main task~\cite{OLAUXNeurIPS2019, AuxLearnICLR2021, MetaBalanceWWW2022}.

\section{Rethinking Task Dominance in MTL}
\label{sec:synthetic}

In this section, we want to answer the following questions: 
(RQ1) How can we quantify the task dominance of a parameter in MTL models?
(RQ2) To what existing do exist MTL approaches tackle the task dominance issue?
(RQ3) How does task dominance impact the training of MTL models?

\subsection{Synthetic Dataset Setting}
To answer the above research questions, we generate a synthetic MTL dataset with two regression tasks, following GradNorm~\cite{chen2018gradnorm}. Specifically, the ground truth for task $k \in \{A, B\}$ is obtained as follows:
\begin{equation}
\setlength\abovedisplayskip{4pt}
\setlength\belowdisplayskip{4pt}
\begin{aligned}
     \mathbf{y}^k(\mathbf{x}) =   \mathbf{W}_1^k \mathbf{x} + \mathbf{W}_2^k \mathbf{x}^2 + \mathbf{W}_3  \mathbf{x}^3  + \mathbf{\epsilon}, &
\label{eq:synthetic}
\end{aligned}
\end{equation}
where $\small \mathbf{W}_1^A$ and $\small \mathbf{W}_2^A$ are constant matrices with elements sampled from $\small \mathcal{N}(1,1)$, $\small \mathbf{W}_1^B$, $\small \mathbf{W}_2^B$ and $\small \mathbf{W}_3$ are constant matrices with elements sampled from $\small \mathcal{N}(10,10)$, and $\small \mathbf{\epsilon}$ is a Gaussian noise sampled from $\small \mathcal{N}(0, 0.1)$. Note that $\small \mathbf{W}_1^B$, $\small \mathbf{W}_2^B$ is roughly ten times as much as $\small \mathbf{W}_1^A$ and $\small \mathbf{W}_2^A$, making the dataset dominated by task $B$. Moreover, $\small \mathbf{x} \in \mathbb{R}^{250}$ and $\small \mathbf{y}^k(\mathbf{x}) \in \mathbb{R}^{100}$. The MTL model is a $4$-layer fully connected neural network with ELU~\cite{elu_iclr2016} as the activation function. We use the RMSProp optimizer for parameter optimization. 

\subsection{(RQ1) How can we quantify the task dominance of parameters in MTL model?}
\label{sec:RQ1}

We quantify the task dominance of a parameter by dividing the accumulative updates from the corresponding task by the overall updates from all tasks for this parameter. Here we choose the exponentially decaying Average of the squared Updates (AU) to measure the accumulative updates, another choice may also be feasible. Note that we mainly focus on the shared parameters rather than the task-specific parameters since the latter is designed to be only updated and therefore dominated by the corresponding task. 

Given a shared parameter $\small \theta_i$, the AU at step $t$ from task $k$ is defined as:
\begin{equation}
\setlength\abovedisplayskip{4pt}
\setlength\belowdisplayskip{4pt}
\small
    \text{AU}(i,t,k) \!\coloneqq\! G\left[\Delta \theta^{2}\right]_{t, i}^{k} \!=\! \gamma G\left[\Delta \theta^2\right]_{t-1,i}^{k} \!+\! (1-\gamma)[\Delta\theta_{t,i}^{k}]^2
    \label{eq:gradientaccumulatedupdate}
\end{equation}
where $\small \Delta \theta_{t,i}^{k}$ denotes the update of parameter $\small \theta_i$ at setp $t$ from task $k$. 
In RMSProp, $\small \Delta \theta_{t,i}^{k} = -\frac{\eta}{\sqrt{G_{t,i}}+\epsilon}g_{t,i}^k$.

The dominance of task $k$ on a parameter $i$ at step $t$ can therefore be defined as the ratio of its AU over the total AU of all tasks, i.e., 
\begin{equation}
\setlength\abovedisplayskip{4pt}
\setlength\belowdisplayskip{4pt}
\small
    \text{rAU}(i,t,k) \coloneqq \frac{\text{AU}(i,t,k)}{\sum_{k'}\text{AU}(i,t,k')}  = \frac{G\left[\Delta \theta^{2}\right]_{t, i}^k}{\sum_{k'} G\left[\Delta \theta^{2}\right]_{t, i}^{k'}}
    \label{eq:ratio_gradientaccumulatedupdate}
\end{equation}
We denote this AU ratio w.r.t. task $k$, or $\text{rAU}(i,t,k)$ in short.

In order to verify whether AU indeed measures the dominance of a specific task, we conduct the following analysis. On the one hand, we pick up the Top-$1\%$ parameters w.r.t. $\text{rAU}(i,T,B)$, i.e., the Top-1\% parameters that are treated as being dominated by task $B$ at the last step $T$ according to our definition. Random Gaussian noises with mean $0$ and various variances of $\{0.01, 0.02, 0.03, 0.04, 0.05\}$ are added to these parameters; on the other hand, we select the Top-$x\%$ ($x \in \{1,5,10,20\}$) w.r.t. $\text{rAU}(i,T,B)$, and add Gaussian noise with a mean of 0 and a variance of 0.01.
We then evaluate the performance of these models on task $A$ and task $B$, and check their performance degradation, compared with the original model. If a parameter is dominated by task $B$, then adding Gaussian noise may lead to a more significant performance deterioration on task $B$ than on task $A$.

\begin{figure}[t]
\begin{center}
\includegraphics[width=.48\linewidth]{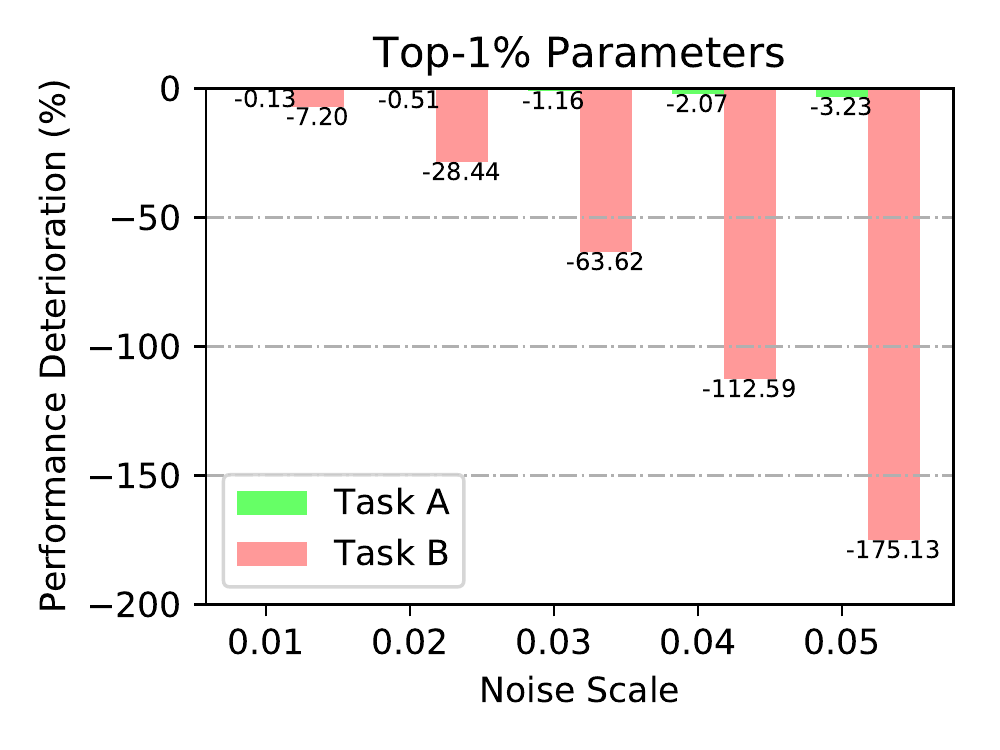}
\includegraphics[width=.48\linewidth]{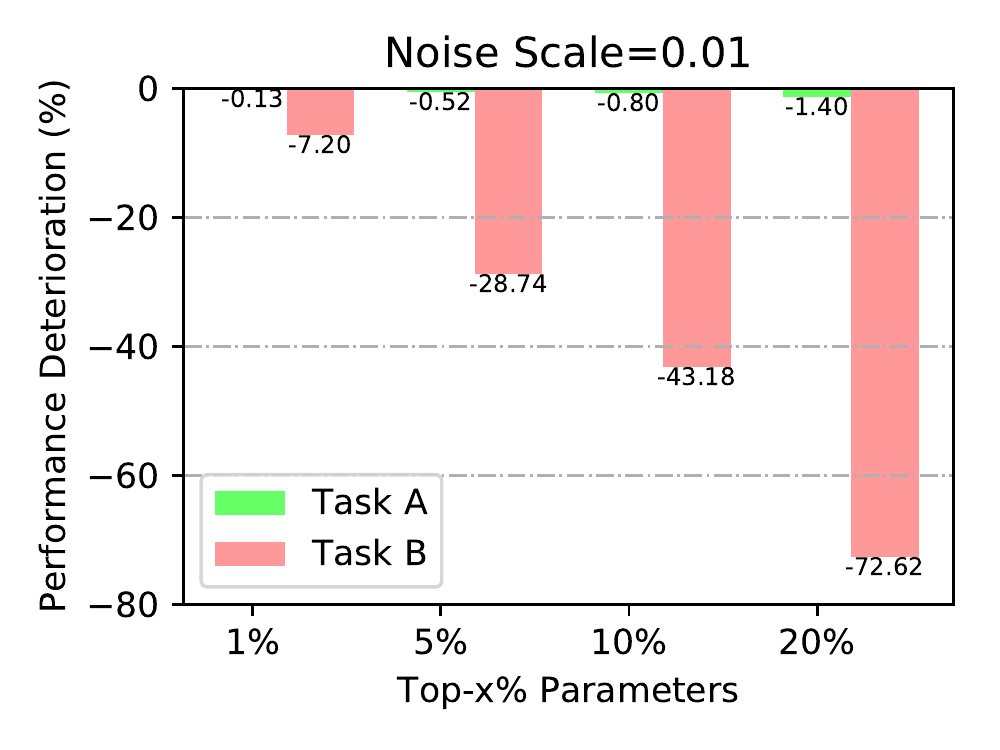}
\end{center}
\caption{Illustration of $\text{rAU}(i,T,B)$ as a metric to measure the task dominance for shared parameters in MTL. 
}
\label{fig:EW_GN_AdaTask_toy_drop}
\end{figure}

We plot the relative {root mean square error} (RMSE) increment of these models in Fig.~\ref{fig:EW_GN_AdaTask_toy_drop}. We observe that \textbf{(1)} {With Gaussian noise added to the Top-1\% parameters w.r.t. $\text{rAU}(i,T,B)$, the loss increases much more on task $B$ than task $A$.} Specifically, the loss is increased by 175.13\% on task $B$, but only 3.23\% on task $A$, with a $\mathcal{N}(0, 0.05)$ Gaussian noise. 
\textbf{(2)} {By adding a $\mathcal{N}(0, 0.01)$ Gaussian noise to the the Top-1\% to the Top-20\% parameters w.r.t. $\text{rAU}(i,T,B)$, the performance deteriorates much more on task $B$ than on task $A$.} With more parameters with high $\text{rAU}(i,T,B)$ polluted by the noise, the performance deteriorates more. For example, the loss is increased by 7.20\% when only the Top-1\% parameters w.r.t. $\text{rAU}(i,T,B)$ is polluted, while it's increased by 72.62\% with the Top-20\% polluted. This concludes that the $\text{rAU}(i,t,k)$ is a reasonable metric to measure the dominance of a task on a shared parameter: a parameter with a high $\text{rAU}(i,t,k)$ for one task becomes much more important for that task, while less important for the remaining tasks.

\subsection{(RQ2) To what extent do existing MTL approaches tackle the task dominance issue?}
\label{sec:RQ2}
\begin{figure*}[th]
\centering
\includegraphics[width=1\linewidth]{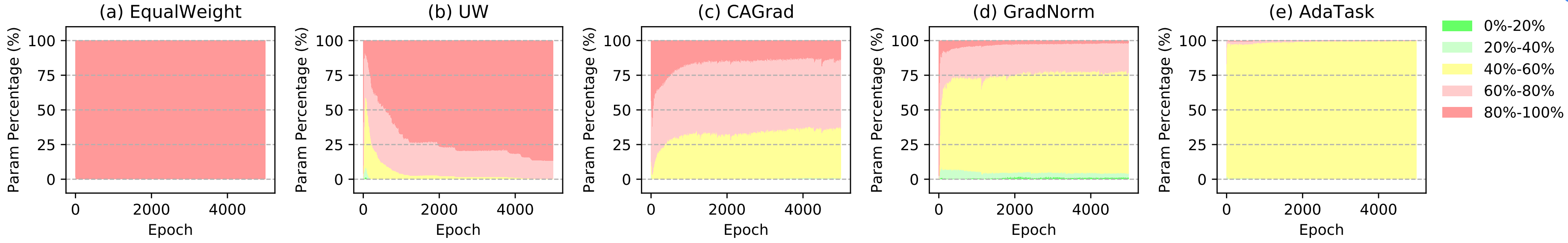}
\caption{$\text{rAU}(i,T,B)$ of all shared parameters on the synthetic dataset for five MTL models: (a) EqualWeight (PCGrad is close to EqualWeight, it was removed due to page limitations.), (b) UW, (c) CAGrad, (d) GradNorm and (e) our AdaTask. The green area denotes the percentage of parameters dominated by task $A$, the red area denotes the percentage of parameters dominated by task $B$, and the yellow area denotes the percentage of balanced parameters.
}
\label{fig:EW_GN_AdaTask_toy_ratio}
\vspace{-10pt}
\end{figure*}

In order to quantify to what extent existing MTL approaches tackle the task dominance issue (taking the dominance of task $B$ as an example here), we calculate the percentage of shared parameters based on their $\text{rAU}(i,T,B)$ using the following pre-defined thresholds: $\{0\%, 20\%, 40\%, 60\%, 80\%, 100\%\}$. %
If $\text{rAU}(i,T,B)$ $\in (80\%,100\%]$, we regard the parameter $i$ as being dominated by task $B$. In addition, if both $\text{rAU}(i,T,A)$ and $\text{rAU}(i,T,B)$ are in $(40\%,60\%]$, we can say that parameter $i$ is roughly balanced. Without loss of generality, we pick up Shared Bottom as the backbone MTL model architecture. Please note that other MTL structures, such as MMOE, PLE, or MTAN are also feasible.  Following~\cite{chen2018gradnorm}, we then pick up UW, CAGrad, and GradNorm as representative MTL balancing methods. We name the vanilla Shared Bottom method as EqualWeight since it corresponds to weighting all tasks equally. 

The percentage of shared parameters of each model w.r.t. $\text{rAU}(i,T,B)$ in different buckets is shown in Fig.~\ref{fig:EW_GN_AdaTask_toy_ratio}. We observe that:
\textbf{(1)} {Almost all shared parameters in EqualWeight are dominated by task $B$}, in the sense that the red area ($\text{rAU}(i,t,B) \in (80\%,100\%]$) is always close to 100\%. This result is because these two methods either use the original gradient directly or use the projected gradient direction, and they have a serious gradient dominance problem when the magnitudes of the gradients of the two tasks are different.
\textbf{(2)} {In UW and CAGrad, some of the parameters are dominated by task $B$.} 
This can be explained as both methods adjust the magnitude of the gradient by weighting the loss or weighting the gradient, thus alleviating the problem of accumulative gradient dominance. However, their weights are designed in terms of learning speed or gradient direction optimization and are not directly related to the magnitude of the gradient, thus only slightly mitigating task dominance.
\textbf{(3)} {In contrast, around 78\% percent of shared parameters in GradNorm are not dominated by any task.}
This is because the weights of the loss in GradNorm are designed to balance the overall gradient magnitude of the task. 

We further split the shared parameters into layers, and calculate the $\text{rAU}(i,t,B)$ for parameters from each layer, respectively. The detailed result is shown in Fig.~\ref{fig:EW_GN_AdaTask_toy_layer_ratio} of Appendix D. 
In EqualWeight, PCGrad and UW, parameters from all layers are dominated by task $B$ in the sense parameters from all layers are with $\text{rAU}(i,t,B) \in (80\%,100\%]$. 
CAGrad and GradNorm are both balanced at some level. Specifically, In GradNorm, only the first layer is balanced well in the sense that around 90\% of the parameters with $\text{rAU}(i,t,B) \in (40\%,60\%]$. However, this percentage drops to around 62\%, 61\%, and 40\%, respectively for the 2nd, 3rd, and 4th layers. 
Such failure in balancing shared parameters from higher layers may be due to that the lower layers learn more general representations while the higher layers learn task-specific representations~\cite{DBLP:conf/nips/YosinskiCBL14}.

\subsection{(RQ3) How does task dominance impact the training of MTL models?}
\label{sec:RQ3}

We have observed that some parameters, especially those in higher layers, are still dominated by task $B$. We then wonder what would happen due to such dominance? Or more specifically, what will happen if the Accumulative Updates (AU) are dominated? 

In modern optimizers, the accumulative gradients are used to adjust the learning rate of each parameter during model training. Then dominance of accumulative gradients may lead to a dominance of learning rate(LR) in MTL. For example, in a MTL setting with only two tasks $A$ and $B$, the accumulative gradients of task $A$ is small, and it would have a large LR if it's trained on its own. However, if task $B$ has larger accumulative gradients, then the overall accumulative gradients in MTL would also be large, leading to a small overall learning rate. With a much smaller learning rate in MTL than in training $A$ alone, task $A$ may be optimized much slower, leading to performance deterioration.

To verify this,  we randomly select multiple parameters from each layer. We then calculate the separate and whole accumulative gradients, as well as the overall LR when training two tasks simultaneously in MTL, as well as the separate LR for each task if it's trained independently. The results are shown in Fig.~\ref{fig:sythetic_learning_rate_dominance}(a) of Appendix D. Each line represents the average learning rate of multiple parameters. It is obvious that the whole LR is dominated by task $B$ in the sense it's very close to the separate LR of task $B$.

\section{Our Proposed Method: AdaTask}
\label{sec:method}

It is observed in the above section existing MTL optimization methods still suffer from task dominance, and such dominance leads to the dominance of learning rate in optimizers. In order to tackle such dominance, in this section, we propose a Task-wise Adaptive learning rate approach, to separate the accumulative gradients of each task for each shared parameter. We name this approach as \emph{AdaTask}.

\noindent
\textbf{Preliminaries}
Adaptive learning rate optimizers adjust the learning rate of each parameter based on some statistics $\small G_{t,i}$ of history gradients w.r.t. this parameter: $\small  \theta_{t+1,i} = \theta_{t,i} -\frac{\eta}{\sqrt{ {G}_{t,i}} + \epsilon} {g}_{t,i}$, where $\eta$ is a hyper-parameter representing the initial global learning rate, $\epsilon$ is a small number used to avoid a zero denominator.
In AdaGrad, $\small G_{t,i}$ is defined as the sum of all history gradient squares, i.e., $\small G_{t,i} = G_{t-1,i} + (g_{t,i})^2$, where $g_{t,i}$ denotes the gradient of this parameter at step $t$. In AdaDelta, Adam or RMSProp, it's defined as the exponential moving average, i.e., $\small G_{t,i} = \gamma G_{t-1,i} + (1-\gamma)(g_{t,i})^2$, where $\gamma$ is a decaying coefficient.

When optimizing shared parameters in MTL models, these statistics are aggregated over all tasks, i.e.:
\begin{equation}
\small
    \begin{split}
        G_{t,i} = G_{t-1,i} + {(g_{t,i})}^2 & \text{  AdaGrad in MTL} \\
        G_{t,i} = \gamma G_{t-1,i} + (1-\gamma){(g_{t,i})}^2 & \text{  RMSProp in MTL} \\
    \end{split}
    \label{eq:xinMTL}
\end{equation}
where $\small g_{t,i} = g_{t-1,i} + \sum {(g_{t,i}^k)}^2$ and $k \in \{1, 2, \ldots, K\}$ represents the $k$-th task.

\noindent
\textbf{AdaGrad, RMSProp with AdaTask.}
As discussed in the above section, the accumulative gradients across tasks in Eq.~\ref{eq:xinMTL} tend to be dominated by some tasks, which in turn leads to the dominance of the adaptive learning rate.
We propose to separate the accumulative gradients of different tasks for each shared parameter. Formally, we maintain task-wise accumulative gradients $G_{t,i}^k$ for parameter $i$ and task $k$ at step $t$. Each such variable only aggregates gradients from the corresponding task, i.e., 
\begin{equation}
\small
    G_{t,i}^k = f(G_{t-1,i}^k, \; (g_{t,i}^k)^2)
    \label{eq:AdaTask_G}
\end{equation}
where $\small f(\cdot)$ denotes the corresponding aggregation function in each method, i.e., average in AdaGrad, and exponentially decaying average in RMSProp. During MTL model training, when optimizing a sample from one task, only the accumulative gradients w.r.t. that task are used to calculate the learning rate. 
Therefore, AdaGrad with AdaTask and RMSProp with AdaTask in MTL can then be formulated as:
\begin{equation*}
\small
    \begin{split}
      \theta_{t+1,i} = \theta_{t,i} - \sum_k & \frac{\eta}{\sqrt{ {G}_{t,i}^k} + \epsilon} {g}_{t,i}^k \\
        G_{t,i}^k \!=\! G_{t-1,i}^k \!+\! {(g_{t,i}^k)}^2  &\text{AdaGrad with AdaTask in MTL} \\
        G_{t,i}^k \!=\! \gamma G_{t-1,i}^k \!+\! (1-\gamma){(g_{t,i}^k)}^2 & \text{RMSProp with AdaTask in MTL} \\
    \end{split}
\end{equation*}

\noindent
\textbf{Adam with AdaTask.}
The Adam~\cite{adam} includes the decayed average of both past gradient squares $G_{t,i}$ as AdaDetla~\cite{ADADELTA} (or RMSProp) and past gradients $m_{t,i}$ as momentum~\cite{momentum1999}. Adam's update rule in MTL is as follows:
\begin{equation}
\small
    \begin{split}
        \theta_{t+1,i} &= \theta_{t,i} - \frac{\eta}{\sqrt{ {G}_{t,i}} + \epsilon} {m}_{t,i} \\
        g_{t,i} &= g_{t-1,i} + \sum {(g_{t,i}^k)}^2   \\
        G_{t,i} &= \gamma_1 G_{t-1,i} + (1-\gamma_1){(g_{t,i})}^2\\
        m_{t,i} &= \gamma_2 m_{t-1,i} + (1-\gamma_2){(g_{t,i})} 
    \end{split}
\end{equation}

When applying AdaTask to Adam, if we only separate the $G_{t,i}$ per task, the accumulation of gradients in $m_{t, i}$ can still be dominated by one or several tasks due to the different magnitudes of $g_{t,i}^k$. In order to prevent task dominance in $m_{t,i}$, we propose to separate it for each task. Adam with AdaTask can be formulated as:
\begin{equation}
\small
    \begin{split}
        \theta_{t+1,i}^k &= \theta_{t,i} - \sum_k \frac{\eta}{\sqrt{ {G}_{t,i}^k} + \epsilon} {m}_{t,i}^k \\
        G_{t,i}^k &= \gamma_1 G_{t-1,i}^k + (1-\gamma_1){(g_{t,i}^k)}^2\\
        m_{t,i}^k &= \gamma_2 m_{t-1,i}^k + (1-\gamma_2){(g_{t,i}^k)} 
    \end{split}
\end{equation}

In Alg.~\ref{alg:xwithoutadatask} / Alg.~\ref{alg:xwithadatask}, we shown the AdaGrad, RMSProp, and Adam without / with AdaTask in MTL, respectively.

\begin{algorithm}[t]
\caption{ {AdaGrad, RMSProp and Adam} in MTL
}
\label{alg:xwithoutadatask}
\small
\small
\begin{algorithmic}[1]
    \STATE{\textbf{Require } $\gamma, \gamma_1, \gamma_2$: Exponential decay factors} 
    
    \STATE{\textbf{Require } $\eta$: Initial learning rate}
    
    \STATE{\textbf{Require } $\epsilon$: A smoothing factor to avoid division by zero}
    
    \STATE{\textbf{Require } $f^k(\theta)$:  Objective function of task $k$, $\forall k \in \{1,2,\ldots,K\}$} 
    
    \STATE{\textbf{Require } Initialize: $G_{0}=0, m_{0}=0$}

    \STATE{}\textbf{For} step $t = 1: T$ \textbf{do}
    
    \STATE{}\ \ \ \  $g_{t,i} = 0$
    
    \STATE{}\ \ \ \  \textcolor{black}{\textbf{For} task $k = 1: K$ \textbf{do}}
    
    \STATE{}\ \ \ \ \ \ \ \  $g_{t,i}^k = \nabla  f_t^k\left(\theta_{t,i}\right)$
    
    \STATE{}\ \ \ \ \ \ \ \  $g_{t,i} = g_{t,i} + g_{t,i}^k$

    \STATE{}\ \ \ \  \textbf{End For}

    \STATE{}\ \ \  \  \textbf{If} AdaGrad 
    
    \STATE{}\ \ \ \ \ \ \ \ \ \ {$G_{t,i}$} = {$G_{t-1,i}$} $+ {g_{t,i}}^2$ 
    
    \STATE{}\ \ \ \ \ \ \ \  \ \ $\Delta\theta_{t,i} =$ {$\frac{\eta}{\sqrt{ {G_{t,i}} } + \epsilon}$}
    $g_{t,i}$

    \STATE{}\ \ \ \ \textbf{Else If} RMSProp 

    \STATE{}\ \ \ \ \ \ \ \ \ \  {$G_{t,i}$} $= \gamma$ {$G_{t-1,i}$} $+ (1-\gamma){g_{t,i}}^2$  
    
    \STATE{}\ \ \ \ \ \ \ \  \ \ $\Delta\theta_{t,i} =$  {$ \frac{\eta}{\sqrt{ {G_{t,i}} } + \epsilon}$} $g_{t,i}$ 
    
    \STATE{}\ \ \  \  \textbf{Else If} Adam 

    \STATE{}\ \ \ \ \ \ \ \ \ \  {$G_{t,i}$} $= \gamma_1$ {$G_{t-1,i}$} $+ (1-\gamma_1){g_{t,i}}^2$ 
    
    \STATE{}\ \ \ \ \ \ \ \ \ \  {$m_{t,i}$} $= \gamma_2$ {$m_{t-1,i}$} $+ (1-\gamma_2){g_{t,i}}$

     \STATE{}\ \ \ \ \ \ \ \ \ \  {$\hat{G}_{t,i}$} = {${G}_{t,i}$} $/ (1-\gamma_1^t)$
 
    \STATE{}\ \ \ \ \ \ \ \ \ \  {$\hat{m}_{t,i}$} = {${m}_{t,i}$} $/ (1-\gamma_2^t)$ 
    
    \STATE{}\ \ \ \ \ \ \ \ \ \  $\Delta\theta_{t,i} =$
    {$\frac{\eta}{\sqrt{ \hat{G}_{t,i}} + \epsilon} \hat{m}_{t,i}$} 
    
    \STATE{}\ \ \ \ \textbf{End If}

    \STATE{}\ \ \ \  $\theta_{t+1,i} = \theta_{t,i} - \Delta \theta_{t,i}$ 

    \STATE{} \textbf{End For}
    
\end{algorithmic}
\end{algorithm}
\begin{algorithm}[t]
\caption{AdaGrad, RMSProp and Adam \textbf{with AdaTask} in MTL
}
\label{alg:xwithadatask}

\begin{algorithmic}[1]
\small
    \STATE{\textbf{Require } $\gamma, \gamma_1, \gamma_2$: Exponential decay factors} 
    
    \STATE{\textbf{Require } $\eta$: Initial learning rate}
    
    \STATE{\textbf{Require } $\epsilon$: A smoothing factor to avoid division by zero}
    
    \STATE{\textbf{Require } $f^k(\theta)$:  Objective function of task $k$, $\forall k \in \{1,2,\ldots,K\}$} 
    
    \STATE{\textbf{Require } Initialize: $G_{0}^k=0, m_{0}^k=0$, $\forall k \in \{1,2,\ldots,K\}$} 
    
    \STATE{}\textbf{For} step $t = 1: T$ \textbf{do}
    
    \STATE{}\ \ \ \  \textbf{If} AdaGrad with AdaTask
    
    \STATE{}\ \ \ \ \ \  \ \ \ \ \textcolor{black}{\textbf{For} task $k = 1: K$ \textbf{do}}
    
    \STATE{}\ \ \ \ \ \ \ \  \ \  \ \  \ \  \ \ $g_{t,i}^k = \nabla  f_t^k\left(\theta_{t,i}\right)$
    
    \STATE{}\ \ \ \ \ \ \ \  \ \  \ \  \ \  \ \   {$\color{black}{G_{t,i}^{{k}}}$} = {$\color{black}{G_{t-1,i}^{{k}}}$} $+ {g_{t,i}^k}^2$ 
    
     \STATE{}\ \ \ \ \ \ \ \  \ \  \ \  \ \  \ \   $\Delta\theta_{t,i}^k =$ {$\frac{\eta}{\sqrt{ \color{black}{G_{t,i}^{{k}}} } + \epsilon}$}
    $g_{t,i}^k$ 
    
    \STATE{}\ \ \ \ \ \  \ \  \ \ \textbf{End For}
    
    \STATE{}\ \ \ \  \textbf{Else If} RMSProp with AdaTask
    
    \STATE{}\ \ \ \ \ \  \ \ \ \ \textcolor{black}{\textbf{For} task $k = 1: K$ \textbf{do}}
    
     \STATE{}\ \ \ \ \ \ \ \  \ \  \ \  \ \  \ \ $g_{t,i}^k = \nabla  f_t^k\left(\theta_{t,i}\right)$
   
    \STATE{}\ \ \ \ \ \ \ \  \ \  \ \ \ \ \ \    {$\color{black}{G_{t,i}^{{k}}}$} $= \gamma$ {$\color{black}{G_{t-1,i}^{{k}}}$} $+ (1-\gamma){g_{t,i}^k}^2$ 
    
     \STATE{}\ \ \ \ \ \ \ \  \ \  \ \ \ \  \ \    $\Delta\theta_{t,i}^k =$  {$ \frac{\eta}{\sqrt{ {\color{black}{G_{t,i}^{{k}}}} } + \epsilon}$} $g_{t,i}^k$  
    
    \STATE{}\ \ \ \ \ \  \ \  \ \ \textbf{End For}
    
    \STATE{}\ \ \ \   \textbf{Else If} Adam with AdaTask
    
    \STATE{}\ \ \ \ \ \  \ \ \ \ \textcolor{black}{\textbf{For} task $k = 1: K$ \textbf{do}}
   
    \STATE{}\ \ \ \ \ \ \ \  \ \  \ \  \ \  \ \ $g_{t,i}^k = \nabla  f_t^k\left(\theta_{t,i}\right)$
   
    \STATE{}\ \ \ \ \ \ \ \  \ \  \ \  \ \  \ \   {$\color{black}{G_{t,i}^{{k}}}$} $= \gamma_1$ {$\color{black}{G_{t-1,i}^{{k}}}$} $+ (1-\gamma_1){g_{t,i}^k}^2$ 

    \STATE{}\ \ \ \ \ \ \ \  \ \ \ \  \ \  \ \    {$\color{black}{m_{t,i}^{{k}}}$} $= \gamma_2$ {$\color{black}{m_{t-1,i}^{{k}}}$} $+ (1-\gamma_2){g_{t,i}^k}$ 

   \STATE{}\ \ \ \ \ \ \ \  \ \  \ \  \ \ \ \    {$\color{black}{\hat{G}_{t,i}^{{k}}}$} = {$\color{black}{{G}_{t,i}^{{k}}}$} $/ (1-\gamma_1^t)$  

    \STATE{}\ \ \ \ \ \ \ \  \ \  \ \  \ \ \ \    {$\color{black}{\hat{m}_{t,i}^{{k}}}$} = {$\color{black}{{m}_{t,i}^{{k}}}$} $/ (1-\gamma_2^t)$   

    \STATE{}\ \ \ \ \ \ \ \  \ \  \ \  \ \ \ \    $\Delta\theta_{t,i}^k =$
    {$\frac{\eta}{\sqrt{ \color{black}{\hat{G}_{t,i}^{{k}}}} + \epsilon} \color{black}{\hat{m}_{t,i}^{{k}}}$} 
    \STATE{}\ \ \ \ \ \  \ \  \ \ \textbf{End For}

    \STATE{}\ \ \ \  \textbf{End If}

    \STATE{}\ \ \ \ $\theta_{t+1,i} = \theta_{t,i} - \sum_k \Delta \theta_{t,i}^k$ 

    \STATE{}\textbf{End For}
    \end{algorithmic}
\end{algorithm}

\noindent
\textbf{Efficient AdaTask.}
In order to separate the Accumulative Gradients in AdaTask, $K\!-\!1$ times more intermediate variables are needed during optimization. This is negligible when there are many shared parameters.
We can reduce the number of variables to store accumulative gradients by grouping parameters by some dimensions, e.g., layers or modules. Without loss of generality, we discuss grouping accumulative gradients by layers here. That is, all parameters in each layer ($\small l \!\! \in \!\! \{1, \! \ldots \!, L\}$) share the same accumulative gradients per task and formalize it as follows:
\begin{equation}
\small 
    \begin{split}
        C_{t, l}^k &=  \gamma C_{t-1, l}^k + (1-\gamma) h \left(  \{ {(g_{t,i}^k)}^2, \forall i \in l{(\theta)}  \} \right),
    \end{split}
\end{equation}
where $\small h(\cdot)$ represents an aggregate function, such as mean or sum operation, which is used to aggregate the gradient of all parameters at the $l$-th layer, $\forall i \in l{(\theta)}$ represents all parameters in the $l$-th shared layer. Therefore, $C_{t, l}^k$ can be regarded as the overall contribution of task $k$ to the shared layer $l$ at step $t$. The accumulative gradient of task $k$ w.r.t parameter $i$ is expressed as follows:
\begin{equation}
\small 
    \begin{split}
        G_{t,i}^k = \frac{C_{t, l(i)}^k}{\sum_k C_{t,l(i)}^{k}} G_{t,i},
        G_{t,i} = \gamma G_{t-1,i} + (1-\gamma){(g_{t,i}^k)}^2, 
    \end{split}
\label{eq:ladatask}
\end{equation}
where $\small l(i)$ which layer parameter $i$ belongs to. Due to limited space, we provide a detailed comparison of AdaTask with the efficient version (LAdaTask) in Appendix A.

\noindent
\textbf{Implementation.}
AdaTask can be implemented in a single backward propagation by separating and passing through the partial derivative of each task's loss.

\section{Experiments}
\label{sec:exp}
In this section, we experimentally verify the effectiveness of the proposed AdaTask. We conduct experiments on a synthetic dataset, and several real-world datasets: the CityScapes dataset in computer vision, and the TikTok and WeChat datasets in recommender systems. For the baseline methods, we choose two loss weighting approaches GradNorm~\cite{chen2018gradnorm}, UW~\cite{uncertaintyweighting}, and two gradient conflict-avoiding approaches  PCGrad~\cite{gradientsurgery} and CAGrad~\cite{cagrad-liu2021conflict}. 
More details of our experiment, including baselines, datasets, and implementation details, are provided in Appendix B. 

We adopt the commonly used \textit{SharedBottom} architecture (mainly based on fully connected layers) and \textit{RMSProp} optimizer in the synthetic and recommendation datasets. In the computer vision dataset, we choose the commonly used \textit{MTAN} architecture (including complex operations such as convolution and attention mechanisms, etc.) and the \textit{Adam} optimizer. Each experiment is repeated over {3 random seeds} and the mean is reported. Validation on different network architectures (such as MMoE and SegNet) is presented in Appendix C. In addition, we provide a comparison of the training times in Appendix E.

\subsection{Performance Evaluation}
\label{sec:exp_performance}
\noindent
\textbf{CityScapes Dataset Results.} For the semantic segmentation and depth estimation tasks on {CityScapes} datasets, we follow MTAN~\cite{2019cvprdwa} and CAGrad~\cite{cagrad-liu2021conflict} to use Absolute Error (Abs Err) and Relative Error (Rel Err) to evaluate the performance of the Depth Estimation task (Task $A$) and use Mean Intersection over Union (mIoU) and Pixel Accuracy (Pix Acc) to evaluate the performance of Semantic Segmentation task (Task $B$). Similar to CAGrad, we also compute the average per-task performance improvement $\Delta p$ of an MTL method $m$ with respect to the EqualWeight baseline $b$: { $\small \! \Delta p\! =\frac{1}{4} \sum_{p}(-1)^{I_{p}}\left(M_{m, p}-M_{b, p}\right) / M_{b, p}$}, where $\small I_p=0$ if a higher value is better for a criterion $M_p$ on performance $p \! \in \! \{\text{Abs Err, Rel Err, mIoU and Pix Acc}\}$ and $1$ otherwise.

As shown in Table~\ref{tab:cityscapes_mtan}, GridSearch delivers higher performance than Equalweight, but manual weighting may require efforts to find the proper weights. 
AdaTask achieved a significant improvement in two metrics of task $A$ (the dominated task in this dataset). Specifically, Abs Err is improved by $15.78\%$ compared to EqualWeight, and Rel Err is improved by $21.59\%$ compared to EqualWeight. Meanwhile, AdaTask achieves a result close to EqualWeight on task $B$. In addition, in terms of the overall performance, AdaTask has achieved an improvement of $1.29\%$ and $2.84\%$ respectively compared with the best baseline (UW) and the second-best baseline (CAGrad), respectively.  
Notably, AdaTask is orthogonal to the methods of modifying gradient direction, such as CAGrad and PCGrad, so it can be further combined with these methods to improve performance. When combined with PCGrad and CAGrad (The implementation details are in Alg.~\ref{alg:xwithadatask_plus_pcgrad} and Alg.~\ref{alg:xwithadatask_plus_cagrad} of Appendix G), the performance can be further improved by 1.18\% and 1.25\%, respectively.  
Finally, by comparing MGDA (which seeks Pareto optimal solutions), we can find that the solutions of GradNorm, UW, PCGrad, CAGrad, AdaTask, and MGDA do not dominate each other, that is, none of them performs better than the other in all metrics.

\begin{table}[t]
\small
    \centering
    \scalebox{0.89}{%
    \begin{tabular}{l|crrr|rc}
     \hline
     & \multicolumn{2}{c}{{Task A}} &\multicolumn{2}{c|}{{Task B}}  & {} \\
    \cmidrule(lr){2-3}\cmidrule(lr){4-5}
     {Method} & Abs Err$\downarrow$ & Rel Err$\downarrow$ & mIoU$\uparrow$ & Pix Acc$\uparrow$ & $\Delta p\uparrow$ \\
      \hline
     EqualWeight   &  0.0152 &  47.00 & 75.01 & 93.40  & 0.00\% \\ \hdashline
     GridSearch   &  0.0132 &  41.88 & 74.13 & 93.16  & 5.65\% \\
     MGDA & 0.0166 &  \underline{33.50}   &  69.42 &  91.22 & 2.40\%  \\
     GradNorm &  {0.0145} & 44.16 & 75.18 & 93.38  &  {2.71\%} \\
     UW   &  {0.0134} & {36.66} & 74.42 & 93.05 &  {8.05\%} \\
     PCGrad $^\dagger$ & 0.0150 & 42.52  & \textbf{75.46} & \textbf{93.49}  & 2.88\%\\
     CAGrad $^\dagger$  &  {0.0141} & 38.40 & \underline{75.30} & \underline{93.48} &  {6.50}\% \\ 
     \hline
     AdaTask &  \underline{0.0128} & {36.85} & 75.02 & 93.40  &  {{9.34\%}} \\  
     \;\;\;\;+ PCGrad & \textbf{0.0127} & 34.46 &  74.76 & 93.25 &  \underline{10.52\%} \\
     \;\;\;\;+ CAGrad & 0.0133 & \textbf{31.50} &  73.17 & 92.88 &  \textbf{10.59\%} \\
     \hline
    \end{tabular}
    }
    \vspace{-0.3cm}
\caption{Performance evaluation on the CityScapes dataset. The $^\dagger$ symbol represents the gradient direction modification methods, and they are orthogonal to our work. Task $A$ is the dominated task, and Task $B$ is the dominant task.}
    \label{tab:cityscapes_mtan}
\vspace{-10pt}
\end{table}

\noindent
\textbf{Synthetic Dataset Results.} For the regression tasks on the synthetic dataset, we follow GradNorm~\citep{chen2018gradnorm} to use Task-Normalized RMSE (Root Mean Square Error) to evaluate the performance of all methods. As shown in Table~\ref{tab:sythetic}, we observe that the proposed AdaTask achieves a huge RMSE improvement in task $A$, with an absolute improvement of 0.14 and 0.04 compared with the EqualWeight and the best baseline (GradNorm), respectively. Such a huge improvement is due to the fact that task $A$ is dominated dramatically but task $B$ and our proposed method resolve such dominance explicitly. The RMSE absolute reduction of task $B$ is 0.03 compared with EqualWeight. 
Task $B$ showed some decline since it already dominates baseline models and is more inclined by them. For the average results of the two tasks, AdaTask achieved the best performance.  
\begin{table}[t]
\begin{center}
\begin{tabular}{l | cc | c}
 \hline
{Method}         & {Task A} $\downarrow$ & {Task B} $\downarrow$ & {Average} $\downarrow$       \\ \hline
EqualWeight                          &0.2172                    &\underline{0.0113}         &0.1142           \\ \hdashline
GradNorm     &\underline{0.1134}        &0.0226                     &\underline{0.0680}           \\  
UW       &0.2032                    &\textbf{0.0100}            &0.1066           \\  
PCGrad        &0.2221                    &0.0155                     &0.1188           \\ 
CAGrad &0.1495                    &0.0140                     &0.0817           \\ 
\hline
AdaTask                        &\textbf{0.0704}           &0.0417                     &\textbf{0.0560}           \\ 
\hline
\end{tabular}
\end{center}
\vspace{-0.3cm}
\caption{Performance on the synthetic dataset. Task $A$ is the dominated task, and task $B$ is the dominant task.
\label{tab:sythetic}
}
\vspace{-18pt}
\end{table}

\begin{table*}[t]
\centering
\begin{tabular}{l | cc | ll}
 \hline
{Method} & {Task A} $\uparrow$ & {Task B} $\uparrow$ & {Average AUC} $\uparrow$  & {Weighted AUC} $\uparrow$\\  \hline
EqualWeight    &0.9173          &\underline{0.7437}    &0.8305 (+0.00\%)          &0.8652 (+0.00\%)        \\  \hdashline
GradNorm~\cite{chen2018gradnorm}       &0.9182         &0.7431     &0.8307 (+0.02\%)         &0.8657 (+0.05\%)        \\ 
PCGrad~\cite{gradientsurgery}      &0.9199         &0.7413     &0.8306 (+0.01\%)         &0.8663 (+0.12\%)        \\
UW~\cite{uncertaintyweighting}       &\underline{0.9237}         &0.7351     &0.8294 (-0.13\%)         &0.8671 (+0.21\%)        \\ 
CAGrad~\cite{cagrad-liu2021conflict}         &0.9185         &\textbf{0.7455}    &\underline{0.8320} (+0.18\%)        &\underline{0.8666} (+0.16\%)         \\ 
 \hline
AdaTask          &\textbf{0.9312}         &0.7399     &\textbf{0.8356} (+0.61\%)         &\textbf{0.8738} (+0.99\%)        \\ 
 \hline
\end{tabular}
\vspace{-0.2cm}
\caption{Performance evaluation on the TikTok. Task $A$ is the dominated task, and task $B$ is the dominant task.
}
\label{tab:tiktok}
\vspace{-5pt}
\end{table*}

\begin{table*}[t]
\centering
\begin{tabular}{l | cccc | ll}
 \hline
{Method} & {Task A}$\uparrow$ & {Task B}$\uparrow$  & {Task C}$\uparrow$  & {Task D}$\uparrow$ & {Average AUC}$\uparrow$ & {Weighted AUC}$\uparrow$  \\
\hline
EqualWeight                   &0.9080         &0.8912      &{0.8881}         &{0.9514}    &0.9097(+0.00\%)                     &0.9160(+0.00\%)        \\  \hdashline
GradNorm~\cite{chen2018gradnorm}                      &0.9408         &0.9072      &0.8840         &0.9462    &0.9196(+1.08\%)         &0.9192(+0.34\%)        \\       
UW~\cite{uncertaintyweighting}             &\textbf{0.9645}         &0.9187      &0.8654         &0.9451    &\underline{0.9234}(+1.51\%)         &0.9179(+0.19\%)        \\ 
PCGrad~\cite{gradientsurgery}       &0.9182         &0.8987      &\underline{0.8942}         &\underline{0.9520}    &0.9158(+0.67\%)         &{0.9206}(+0.50\%)        \\ 
CAGrad~\cite{cagrad-liu2021conflict}   &0.9462         &\underline{0.9313}         &\textbf{0.9040}         &\textbf{0.9538}   &0.9338(+2.65\%)        &\textbf{0.9336}(+1.91\%)         \\ 
 \hline
AdaTask                      &\underline{0.9643}         &\textbf{0.9391}      &0.8920         &0.9480    &\textbf{0.9359}(+2.87\%)         &\underline{0.9311}(+1.63\%)        \\ 
 \hline
\end{tabular}
\vspace{-0.2cm}
\caption{Performance evaluation on the WeChat. Task $A$, $B$, and $C$ are the dominated tasks, and task $D$ is the dominant task.}
\label{tab:weixin}
\vspace{-13pt}
\end{table*}

\noindent
\textbf{TikTok Dataset Results.} For the MTL classification tasks on the {TikTok} dataset, we evaluate Area under the ROC curve (AUC)~\cite{AUC} of each task, as well as the average and weighed AUC, with weights specified by the provider. As shown in Table~\ref{tab:tiktok}, similar to the observation on the synthetic dataset, AdaTask gets a significant AUC lift on the dominated task ($A$ in this case). The improvement on task $A$ is 1.51\% and 0.81\%, compared with EqualWeight and the best baseline (CAGrad). While on task $B$, there is a slight performance drop of 0.51\%, compared with EqualWeight. Such performance drop may be due to a strong preference for task $B$ in these baseline methods. Compared with EqualWeight, AdaTask improves the Average AUC and Weighted AUC by 0.61\% and 0.99\%, respectively. While against the best baseline model, AdaTask improves the Average AUC and Weighted AUC by 0.59\% (against GradNorm), 0.6\% (against PCGrad), and 0.43\% (against CAGrad).

\noindent
\textbf{WeChat Dataset Results.} For the {WeChat} dataset, similar to Tiktok, we used average AUC and weighted AUC to measure the overall performance. As shown in Table~\ref{tab:weixin}, we observed a similar phenomenon with the above three datasets, that is, AdaTask improved significantly in the dominated tasks (Task $A$, $B$, and $C$), while its performance decreased slightly in the dominate task (Task $D$). Compared to the best baseline method, i.e., CAGrad, AdaTask slightly outperformed it by $0.22\%$ on average AUC. However, w.r.t. the weighted AUC, CAGrad outperforms AdaTask by $0.28\%$. This is because AdaTask usually performs worse on the dominant task, but in the WeChat dataset the dominant one, i.e., Task D,  gets a high weight (The weights for task $A$, $B$, $C$, and $D$ is 1:2:3:4, please refer to Appendix B for details).

\subsection{Overall Results} 
\noindent
\textbf{Multi-task overall performance perspective.} Experiments on four datasets demonstrate that our proposed AdaTask model achieves state-of-the-art performance w.r.t. Average and Weighted metrics. Such improvement mainly comes from the dominated task, which is task $A$ in Synthetic, CityScapes and TikTok cases, task $A$, $B$, and $C$ in WeChat case. 
Compared with baselines, our results show a certain decrease in the dominant tasks. 
This is because baseline methods are heavily skewed toward the dominant task, resulting in better performance for the dominant task. However, our method restricts the shared parameters to be skewed to the dominant task to a certain extent, so the performance of the dominant task will be slightly degraded, but we achieve the overall optimal performance.

\noindent
\textbf{Multi-objective Pareto solution perspective.}
In addition to overall performance, we are also interested in reviewing these results from a MOO perspective.
According to the definition of the Pareto stationary solution, solution $\theta^{*}$ is said to belong to the Pareto stationary solution set if there is no solution $\theta$ that outperforms $\theta^{*}$ in all tasks. So we can consider that the solution found by AdaTask also belongs to Pareto stationary solution because there is no method (such as MGDA, GradNorm, UW, PCGrad, or CAGrad) that outperforms AdaTask on all task metrics. We provide a more detailed comparison and discussion of the MOO approach in Appendix C. 

\subsection{Study on Task Dominance}
\label{sec:exp_analysis}

\noindent
\textbf{Task Dominance on AU.} In this part, we study whether the proposed AdaTask approach indeed resolves the task dominance w.r.t. AU. We calculate the percentage of shared parameters from all layers w.r.t. their $\text{rAU}(i,T,B)$ using some pre-defined thresholds: \{0\%, 20\%, 40\%, 60\%, 80\%, 100\%\}.
The result of AdaTask is shown in Fig.~\ref{fig:EW_GN_AdaTask_toy_ratio}(e) for the synthetic dataset. Very few parameters in AdaTask are dominated by task $B$ (yellow area in (e)). We then further split the parameters into four layers in the synthetic dataset. The results are shown in Fig.~\ref{fig:EW_GN_AdaTask_toy_layer_ratio}. The proposed AdaTask achieves a decent balance of AU: after the first several epochs, more than 98\% of shared parameters in all four layers have $\text{rAU}(i,T,B)$ in (40\%, 60\%]. 
We also show that task dominance is a real problem in the real-world dataset (CityScapes). As shown in Fig.~\ref{fig:cityscapes_au_dominance} of Appendix D, we calculate the AU of EqualWeight(using MTAN) and GradNorm(using MTAN) on the CityScapes dataset and do observe that 99\% and 95\% of shared parameters in these two methods are dominated by task $B$. After applying AdaTask(on MTAN), only 7\% of parameters are slightly dominated by task $B$, while 93\% of shared parameters are balanced well. It shows that AdaTask solves the task dominance issue well.

\noindent
\textbf{Task Dominance on Learning Rate.} In this part, we study the task dominance of the learning rate in AdaTask. 
The learning rate of the baseline methods dominates, which we analyze in Task-dominance analysis section and our AdaTask is shown in Fig.~\ref{fig:sythetic_learning_rate_dominance}(b) of Appendix D.
There is no such learning rate dominance problem in our proposed AdaTask since we separate learning rates between tasks for each parameter. Therefore a parameter can be optimized toward each task's objective based on its own learning rate, without any intervention from other tasks.

\section{Conclusion and Future Works}
\label{sec:conclusion}
In this work, we quantify the dominance of parameter training and demonstrate that such dominance still occurs in existing works and it leads to a serious optimization problem using adaptive learning rate methods for optimization. We propose a Task-wise Adaptive Learning Rate Method, named AdaTask, to use task-specific accumulative gradients when adjusting the learning rate of each parameter. Comprehensive experiments on synthetic and three real-world datasets demonstrate that AdaTask achieves comparable performance with state-of-the-art MTL approaches.

There are several potential directions for future research. First, when AdaTask significantly improves the performance of dominated task, it could damage the performance of the dominant task, so further research is needed to ensure the performance of the dominant task. Second, deep multi-task learning is a highly nonconvex problem, making convergence analysis quite challenging. Therefore, we expect to perform a theoretical convergence analysis for AdaTask in the future.

\section{Acknowledgement}
This work is partially supported by the Major Science and Technology Innovation 2030 “Brain Science and Brain-like Research” key project (No. 2021ZD0201405), the National Natural Science Foundation of China under Grant (No. 62032013, 61972078), and the Fundamental Research Funds for the Central Universities under Grant No. N2217004.

\bibliography{aaai23}

\newpage
\appendix


\section{Appendix A: Efficient AdaTask}
\label{sec:EfficientAdaTask}
In this section, we will discuss and verify the effect of LAdaTask in more detail. Specifically, the number of parameters, including intermediate variables, and the communication cost for each method is summarized in Table~\ref{tab:complexity}.
For the number of parameters or intermediate variables, besides the $N$ parameters, RMSProp (as well as other optimizers) needs to maintain $N$ accumulative gradients $G$, one for each parameter. Parameter wise task-aware adaptive learning rate methods, i.e., AdaTask, as defined in Alg.~\ref{alg:xwithadatask}, need to maintain $NK$ accumulative gradients $G^k$ during model training, and there is $NK$ additional communication cost to transfer these accumulative gradients from workers to parameter servers. For the Layer wise task-aware adaptive learning rate methods, i.e., LAdaTask as defined in Eq.~\ref{eq:ladatask}, only $LK$ accumulative gradients need to be maintained. 

\begin{table}[h]
\small
  \centering
  \begin{tabular}{c|c|c}
    \hline
    Method & \#Intermediate Variables & \#Communication Cost  \\
    \hline
    RMSProp & $O(N)$    & $O(N)$ \\
    AdaTask & $O(NK)$   & $O(NK)$  \\
    LAdaTask & $O(N+LK)$ & $O(N+LK)$ \\
    \hline
    \end{tabular}
\vspace{-0.3cm}
\caption{The number of parameters and communication cost of each method. Here AdaTask, and LAdaTask denote the Parameter-wise and Layer-wise Task-aware Adaptive Learning Rate methods, respectively.}
\label{tab:complexity}
\vspace{-20pt}
\end{table}

\begin{table}[h]
\small
\centering
\scalebox{0.83}{
\begin{tabular}{l l  cc  cc}
 \hline 
{Dataset} &{Method} & {Task A}$\downarrow$ & {Task B}$\downarrow$ & {Average}$\downarrow$  & {Weighted}
\\ \hline
\multirow{2}{*}{Synthetic} 
&EqualWeight          &0.2172                    &0.0113         &0.1142  &   - \\ 
&AdaTask         &0.0704        &0.0417        &0.0560     &   -   \\ 
&LAdaTask           &0.0989        &0.0285        &0.0637     &   -    \\ 
\hline  \hline 
{Dataset} &{Method} & {Task A}$\uparrow$ & {Task B}$\uparrow$ & {Average}$\uparrow$  & {Weighted}$\uparrow$
\\ \hline
\multirow{2}{*}{TikTok}   
&EqualWeight        &0.9173     &0.7437    &0.8305        &0.8652  \\
&AdaTask           &0.9312     &0.7399    &0.8356        &0.8738         \\ 
&LAdaTask           &0.9196          &0.7446   &0.8321            &0.8671          \\ 
  \hline
\end{tabular}
}
\vspace{-0.3cm}
\caption{Performance evaluation of efficient AdaTask (i.e., LAdaTask) on the synthetic dataset (Lower Better) and TikTok dataset (Higher Better). Task $A$ is the dominated task, and task $B$ is the dominant task.}
\label{tab:sythetic_and_tiktok_efficient}
\vspace{-12pt}
\end{table}

\begin{table*}[t]
\centering
\begin{tabular}{l | cccc | ll}
 \hline
{Method} & {Task A} $\uparrow$ & {Task B} $\uparrow$  & {Task C} $\uparrow$  & {Task D} $\uparrow$ & {Average AUC} $\uparrow$ & {Weighted AUC} $\uparrow$  \\
\hline
EqualWeight (SharedBottom)                   &0.9080         &0.8912      &{0.8881}         &{0.9514}     &0.9097 (+0.00\%)                     &0.9160 (+0.00\%)        \\ 
AdaTask (SharedBottom)                       &{0.9643}        &{0.9391}      &0.8920         &0.9480    &{0.9359} (+2.87\%)         &{0.9311} (+1.63\%)        \\ 
 \hdashline
 EqualWeight (MMoE: 4 experts) &0.9126         &0.9052      &0.8898         &0.9512    &0.9147 (+0.00\%)                     &0.9096 (+0.00\%)        \\ 
AdaTask (MMoE: 4 experts)   &0.9568         &0.9263      &0.8949         &0.9499    &0.9320 (+1.89\%)                     &0.9345  (+2.73\%) \\
 \hline
\end{tabular}
\caption{Performance evaluation of different architectures (SharedBottom v.s. MMoE) on the WeChat dataset. Task $A$, $B$, and $C$ are the dominated tasks, and task $B$ is the dominant task.
}
\label{tab:weixin_sb_mmoe}
\vspace{-15pt}
\end{table*}

We conducted experiments on Synthetic and TikTok datasets to evaluate AdaTask and LAdaTask, and the results are shown in Table~\ref{tab:sythetic_and_tiktok_efficient}. We observe that LAdaTask is better than EqualWeight on the dominated task (Task $A$), but the improvement is not as significant as AdaTask. Specifically, on the synthetic dataset, LAdaTask achieves an average Task-Normalized RMSE of 0.0637 (lower better), while EqualWeight and AdaTask are 0.1142 and 0.0560, respectively. On the TikTok dataset, the average AUC of LAdaTask is 0.8321 (weighted AUC is 0.8671), and the higher the two evaluation metrics, the better the performance. The EqualWeight and AdaTask are 0.8305(0.8652) and 0.8356(0.8738), respectively. 

\begin{figure}[h]
    \centering
    \setlength{\abovecaptionskip}{0.cm}
    \includegraphics[width=1\linewidth]{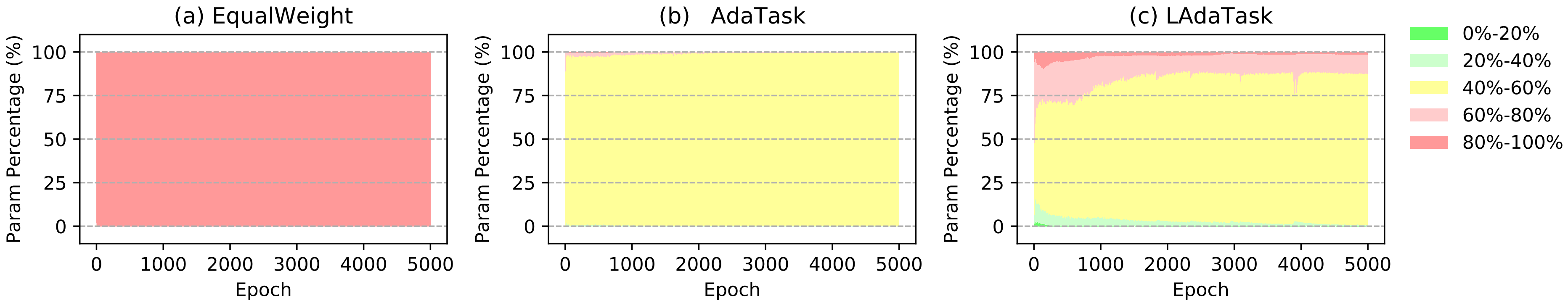}
    \caption{$\text{rAU}(i,T,B)$ of shared parameters on the synthetic dataset for EqualWeight(RMSprop), AdaTask, and LAdaTask methods. 
    }
\label{fig:LAdaTask_toy_layer_ratio}
\vspace{-10pt}
\end{figure}

In addition, we also study LAdaTask's dominance in the Synthetic dataset. As shown in Fig.~\ref{fig:LAdaTask_toy_layer_ratio}(b) and (c), about $85\%$ of LAdaTask's parameters are not dominated by any task, while almost all parameters in AdaTask are not dominated. However, in EqualWeight, almost all parameters are dominated by task $B$ (Fig.~\ref{fig:LAdaTask_toy_layer_ratio}(a)). This shows that LAdaTask can alleviate the task dominance problem to a certain extent. 

\section{Appendix B: Experiment Details}
\label{sec:ExperimentDetails}
This section describes the details of our experimental setting, including baseline methods, datasets, and parameter settings.

\subsection{Baselines}
\label{sec:Baselines}
We present the performance mainly comparison of AdaTask with the following MTL approaches:
\begin{itemize}
    \item \textbf{EqualWeight}~\cite{shared_bottom}: A basic MTL model that learns shared bottom across tasks as well as task-specific towers.
    \item \textbf{GradNorm}~\cite{chen2018gradnorm}: A MTL optimization approach that automatically balances training in deep multitask models by normalizing gradients across tasks.
    \item \textbf{Uncertainty Weighting} (UW)~\cite{uncertaintyweighting}: A MTL optimization approach that weighs multiple loss functions by considering the homoscedastic uncertainty of each task.
    \item \textbf{PCGrad/Gradient Surgery}~\cite{gradientsurgery}: A MTL optimization technique that projects the gradient of a task onto the normal plane of the gradient of any other task to avoid gradient conflicts.
    \item \textbf{CAGrad}~\cite{cagrad-liu2021conflict} A MTL optimization technique that seeks a gradient update direction in the mean gradient direction neighborhood, to maximize the sub-task with the smallest lift in all tasks. 
\end{itemize}

\subsection{Datasets}
\label{sec:exp_dataset}

We conducted experiments on three large-scale real-world datasets, namely CityScapes, TikTok, and WeChat. 
\begin{itemize}
    \item \textbf{CityScapes} dataset is a multitask learning benchmark for depth estimation (Task $A$) and semantic segmentation (Task $B$). Cityscapes have 5,000 high-resolution images of street scenes from 50 different cities\footnote{\footnotesize \url{https://www.cityscapes-dataset.com/}}. In this paper, we use the dataset provided by MTAN\footnote{\footnotesize \url{https://github.com/lorenmt/mtan}}. 
    \item \textbf{TikTok} dataset comes from the ICME 2019 Short Video Content Understanding and Recommendation Competition\footnote{\footnotesize \url{ https://www.biendata.xyz/competition/icmechallenge2019/}}. TikTok consists of two tasks (Like, Finish), named Task $A$ and Task $B$, with official weights of 7:3. 
    The complete dataset contains 19,622,340 samples and 4,691,488 features.
    Since the dataset does not provide timing information in logs, we randomly divided it into training, validation and test sets with a ratio of 8:1:1.   
    \item \textbf{WeChat} dataset is from the 2021 WeChat Big Data Challenge\footnote{\footnotesize \url{https://algo.weixin.qq.com/2021/intro}}.
    It consists of four tasks (named Task $A$-$D$), with official weights of 1:2:3:4. It consists of 8,373,653 interaction data samples across 14 days. To simulate the real-world setting, we split the WeChat data by dates, i.e., the first 12 days as the training set, the next one day as the validation set, and the last day as the testing set.
\end{itemize}

\subsection{Implementation Details}
\label{sec:impl}

To evaluate the methods on the \textbf{CityScapes} dataset, we follow the same setting as MTAN~\cite{2019cvprdwa} and CAGrad~\cite{cagrad-liu2021conflict}. Specifically, we use the MTAN~\cite{2019cvprdwa} as the backbone. 
After combining various baseline methods into the backbone,  
we use the Adam~\cite{adam} optimizer with an initial learning rate of $0.0001$ to train $200$ epochs, and halve the learning rate at the $100$-th epoch. The batch size is set to $8$. As MTAN and CAGrad, we do not create a separate validation set, and report the average performance of the last $10$ epochs of the test set.

To evaluate the methods on \textbf{TikTok and WeChat} datasets, we chose Shared Bottom as the backbone, which is widely used in multi-task recommendation systems~\cite{ESMM,AITMKDD2021,MetaBalanceWWW2022}. 
The Shared Bottom consists of an embedding layer and two hidden layers. In addition, each task has a private tower composed of three layers of MLPs with ELU, with the number of neurons as [128, 64, 32], respectively. The embedding size is set as $10$ for all methods on both datasets. We used RMSprop with an initial learning rate of 0.001 and a  batch size of $4,096$. 
For TikTok and WeChat datasets, we implement low-frequency filtering for the original features, so that features that occur less than ten times are uniformly set to ‘Other’. In addition, we use the early-stop strategy and end the training when the result does not improve on the validation set for five consecutive epochs.

In all datasets, we use three random seeds for all methods and report average results. For all baseline methods, we use the parameter search range recommended in the original paper~\citep{chen2018gradnorm,uncertaintyweighting,gradientsurgery,cagrad-liu2021conflict} for hyper-parameters tuning.
All experiments are conducted on an NVIDIA V100 GPU with 16 GB memory.

\section{Appendix C: Compared with MOO methods and More Model Architectures\&Optimizers}
\label{sec:additional_experiments}
In this section, we conduct experiments in the following aspects: 1) Performance comparison and discussion between AdaTask and MOO methods; 2) Validation of AdaTask on more architectures and optimizers.

\subsection{Comparison with MOO methods}
\begin{table}[t]
\small
    \centering
    \scalebox{0.8}{%
    \begin{tabular}{lccccr}
     \hline
     & \multicolumn{2}{c}{{Task A}} &\multicolumn{2}{c}{{Task B}}  & {Rel Avg} \\
    \cmidrule(lr){2-3}\cmidrule(lr){4-5}
     {Method} & AbsErr$\downarrow$ & RelErr$\downarrow$ & mIoU$\uparrow$ & PixAcc$\uparrow$ & $\Delta p\uparrow$ \\
      \hline
     MGDA  & 0.0166 &  \textbf{33.50}   &  69.42 &  91.22 & -13.26\%  \\
     ParetoMTL(0.25, 0.75) & 0.0154 &  38.89   &  74.97 &  \textbf{93.52} & -6.56\%  \\
     ParetoMTL(0.75, 0.25) & 0.0149 &  36.30   &  \textbf{75.18} & 93.45 & -3.66\%  \\
     AdaTask &  \textbf{0.0128} & {36.85} & 75.02 & 93.40  &  {{0.00\%}} \\ 
     \hline
    \end{tabular}
    }
\caption{Performance evaluation of MOO methods (MGDA and ParetoMTL) on the CityScapes dataset. Task $A$ is the dominated task, and task $B$ is the dominant task.}
\label{tab:cityscapes_paretomtl}
\vspace{-10pt}
\end{table}

Recent advances formulate MTL as a Multi-Objective Optimization (MOO) problem~\cite{mtlasmooSenerK18,lin2019paretomtl}, and these approaches achieve a Pareto stationary solution, or a set of Pareto solutions with various trade-off preferences. In this section, we compare AdaTask with ParetoMTL~\cite{lin2019paretomtl}, a representative MOO approach to MTL.
As shown in Table~\ref{tab:cityscapes_paretomtl}, we compare our AdaTask with MGDA~\cite{mtlasmooSenerK18} and ParetoMTL~\cite{lin2019paretomtl} with different preferences.
Specifically, ParetoMTL(0.75, 0.25) and ParetoMTL(0.25, 0.75) represent the Pareto solution inclined to task $A$ and $B$, respectively. We have the following observations.
From the perspective of overall performance (i.e., $\Delta p$), AdaTask is better than Pareto MTL with either (0.25, 0.75) or (0.75, 0.25) preference. From the MOO perspective, AdaTask also achieves a Pareto solution in the sense that Pareto MTL with various preferences doesn't dominate AdaTask.

\subsection{More Model Architectures and Optimizers}
In the main text, we apply the MTAN~\cite{2019cvprdwa} model on the CityScapes dataset and the ShareBottom~\cite{shared_bottom} model on the WeChat dataset.
In this subsection, we apply AdaTask to more multi-task network architectures and optimizers. Specifically, we use the SegNet~\cite{badrinarayanan2017segnet} on the CityScapes dataset, and the MMoE~\cite{mmoe} on the WeChat dataset. In addition, we also tested the RMSProp optimizer on the CityScapes dataset. The results are presented in Table~\ref{tab:weixin_sb_mmoe} and Table~\ref{tab:cityscapes_segnet_mtan}.

On the WeChat dataset, we used average AUC and weighted AUC to
measure the overall performance. As shown in Table~\ref{tab:weixin_sb_mmoe}, by applying AdaTask on MMoE, we found that it could further improve the average AUC by +1.89\% compared with applying EqualWeight on MMoE. In addition, on the MMoE model, AdaTask improves the weighted AUC by $2.73\%$ compared to EqualWeight. Similar to MTAN, MMoE can also allocate capacity among tasks via attention. 

Similar to CAGrad~\cite{cagrad-liu2021conflict}, we compute the average per-task performance improvement $\Delta p$ of AdaTask to the EqualWeight baseline on the CityScapes dataset. As shown in Table~\ref{tab:cityscapes_segnet_mtan}, we found that by applying AdaTask on SegNet, overall performance resulted in a $23.95\%$ improvement. 
We also observe that MTAN outperforms SegNet in all task evaluation metrics (i.e., Abs Err, Rel Err, mIoU, and Pix Acc), because it designs task-specific attention mechanisms to share the network backbone flexibly.

Finally, we add the experiments of RMSprop with AdaTask on the CityScapes dataset. We observe that it brings a 6.84\% average improvement compared to the traditional RMSProp (EqualWeight baseline). These results are consistent with the one using Adam(9.34\% average improvement) and validate the effectiveness of AdaTask for different optimizers on the same dataset. 

\begin{table*}[h]
    \centering
    \begin{tabular}{lccccr}
     \hline
     & \multicolumn{2}{c}{{Task A}} &\multicolumn{2}{c}{{Task B}}  & {Rel Avg} \\
    \cmidrule(lr){2-3}\cmidrule(lr){4-5}
     {Method} & Abs Err$\downarrow$ & Rel Err$\downarrow$ & mIoU$\uparrow$ & Pix Acc$\uparrow$ & $\Delta p\uparrow$ \\
      \hline
     EqualWeight (SegNet model, Adam optimizer)   &  {0.0198} & 121.81 & 70.23 & 91.92 & +0.00\%   \\ 
     AdaTask (SegNet model, Adam optimizer) & {0.0134} & 45.53 & 70.76 & 92.03   &  {+{23.95\%}} \\ 
     \hdashline
     EqualWeight (MTAN model, Adam optimizer)   &  0.0152 &  47.00 & 75.01 & 93.40  & +0.00\% \\ 
     AdaTask (MTAN model, Adam optimizer) &  0.0128 & {36.85} & 75.02 & 93.40  &  {+{9.34\%}} \\ 
     \hdashline
     EqualWeight (MTAN model, RMSProp optimizer)   &  0.0152 &  44.63 & 74.86 & 93.43  & +0.00\% \\ 
     AdaTask (MTAN model, RMSProp optimizer) &  0.0132 & {37.46} & 74.05 & 93.02  &  {+{6.84\%}} \\ 
     \hline
    \end{tabular}
\caption{Performance evaluation of different architectures (SegNet v.s. MTAN) on the CityScapes dataset. Task $A$ is the dominated task, and task $B$ is the dominant task.}
\label{tab:cityscapes_segnet_mtan}
\end{table*}

\begin{figure}[th]
    \centering
    \setlength{\abovecaptionskip}{0.cm}
    \includegraphics[width=\linewidth]{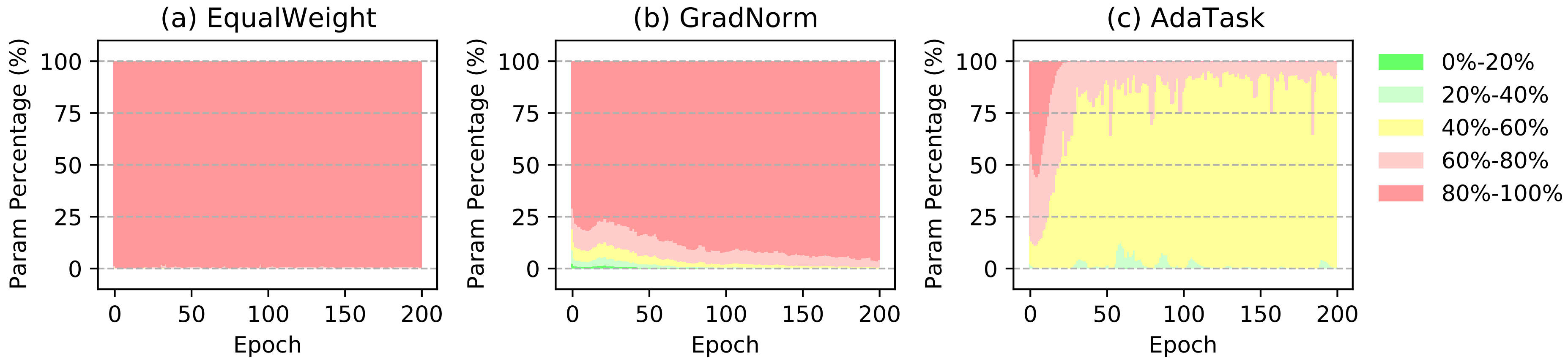}
    \caption{$\text{rAU}(i,T,B)$ of shared parameters on the CityScapes dataset for EqualWeight, GradNorm, and AdaTask methods. 
    }
\label{fig:cityscapes_au_dominance}
\vspace{-15pt}
\end{figure}

\begin{figure}[th]
\vspace{-10pt}
    \centering
    \setlength{\abovecaptionskip}{0.cm}
    \includegraphics[width=\linewidth]{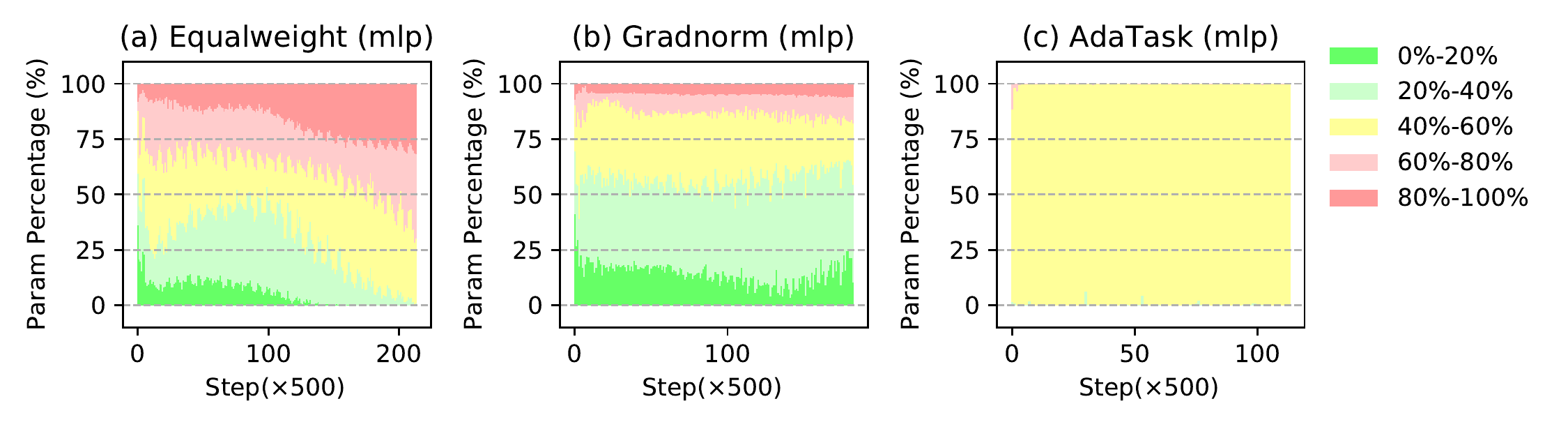}
    \caption{$\text{rAU}(i,T,B)$ of shared parameters (MLP layer) on the TikTok dataset for EqualWeight, GradNorm, and AdaTask methods. 
    }
\label{fig:tiktok_au_dominance_mlp}
\vspace{-15pt}
\end{figure}

\begin{figure}[th]
\vspace{-10pt}
    \centering
    \setlength{\abovecaptionskip}{0.cm}
    \includegraphics[width=\linewidth]{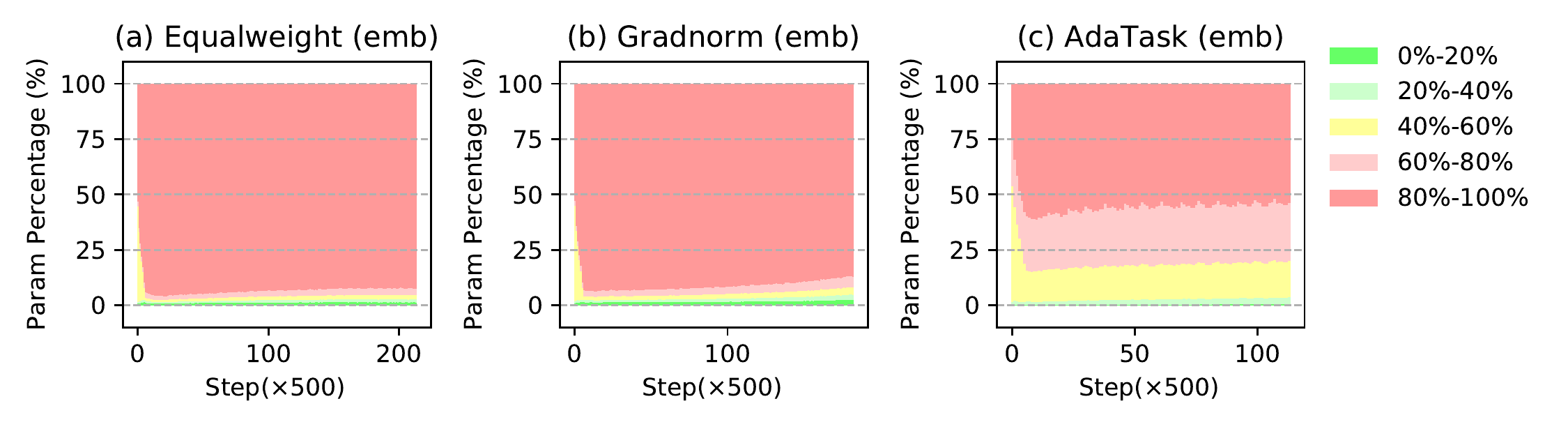}
    \caption{$\text{rAU}(i,T,B)$ of shared parameters (Embedding layer) on the TikTok dataset for EqualWeight, GradNorm, and AdaTask methods. 
    }
\label{fig:tiktok_au_dominance_emb}
\vspace{-15pt}
\end{figure}

\begin{figure*}[th]
    \setlength{\abovecaptionskip}{0.cm}
    \includegraphics[width=\linewidth]{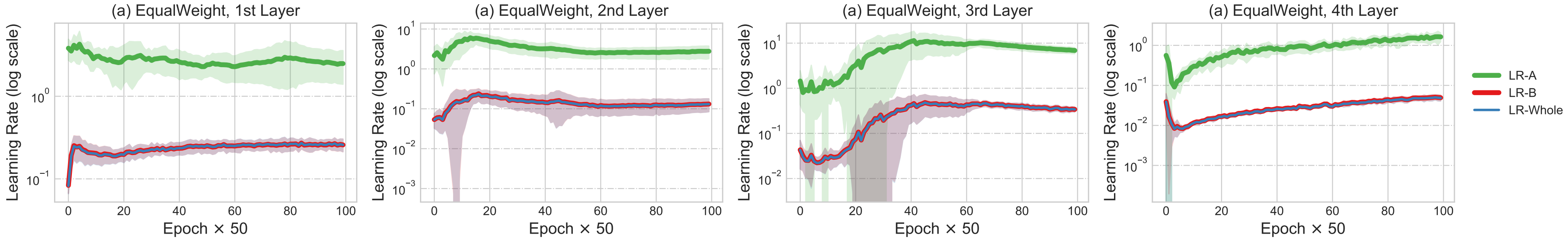}
    \includegraphics[width=0.99\linewidth]{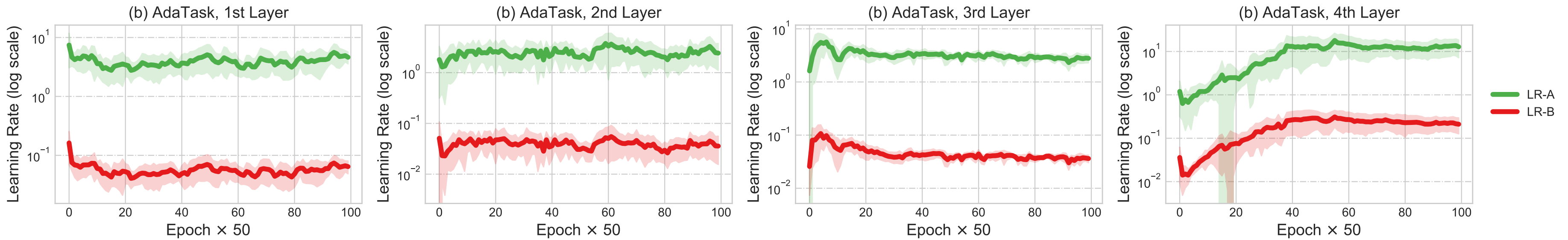}
    \caption{Illustration of learning rate dominance of randomly sampled parameters from the 1st to 4th layer for two MTL methods: (a) EqualWeight and (b) our AdaTask.
    The learning rate of Task $A$ (i.e., \texttt{LR-A}) or Task $B$ (i.e., \texttt{LR-B}) is based on Task $A$'s or Task $B$'s own accumulative gradient, respectively. The whole learning rate (i.e., \texttt{LR-Whole}) is based on the sum of accumulative gradients from both tasks.
    }
    \label{fig:sythetic_learning_rate_dominance}
\vspace{-10pt}
\end{figure*}

\begin{algorithm}[t]
\caption{{CAGrad(Adam) with AdaTask} 
}
\label{alg:xwithadatask_plus_cagrad}
\small
\begin{algorithmic}[1]
\STATE{\textbf{Require } Model parameters $\theta$, a constant $c \in [0,1)$}
    
    \STATE{} \textbf{For} step $t \in \{1,2,\ldots,T\}$ \ \textbf{do}
    
    \STATE{} \ \ $(\textbf{g}_t^k)^\text{PC} \gets \textbf{g}_t^k$ \ $\forall k$
     
    \STATE{}  \ \ $g_0 = \frac{1}{K} \sum_k g^k_t$
    
     \STATE{}  \ \ $\phi = c^2 ||g_0||^2$
     
     \STATE{}  \ \ Solve $\min _{w \in \mathcal{W}} F(w):=g_{w}^{\top} g_{0}+\sqrt{\phi}\left\|g_{w}\right\|$, where $g_{w}=\frac{1}{K} \sum_k w^k g^k$
     
     \STATE{} \ \  \textbf{For} task $k \in \{1,2,\ldots,K\}$ \ \textbf{do}
     
     \STATE{} \ \ \ \   $(\textbf{g}_t^k)^\text{CA} \gets \frac{1}{K} (1 + \frac{\phi^{0.5}}{||g_w||} w^k ) g^k
     $ 
    
    \STATE{} \ \ \ \  $\Delta \theta^k_{t}$ {$ \gets  -\frac{\eta}{\sqrt{\hat{\textbf{G}}^k_{t}} + \epsilon} \hat{\textbf{m}}^k_{t} $}  
        
    \STATE{}\ \ \ \   \text{where } 
    
    \STATE{}\ \ \ \  \ \   {$\hat{\textbf{G}}^k_{t}$} = {${\textbf{G}}^k_{t}$} $/ (1-\gamma_1^t)$, \;\; {$\hat{\textbf{m}}^k_{t}$} = {${\textbf{m}}^k_{t}$} $/ (1-\gamma_2^t)$ 

    \STATE{}\ \ \ \ \ \  
    {$\textbf{G}^k_{t}$} $= \gamma_1$ {$\textbf{G}^k_{t-1}$} $+ (1-\gamma_1)
    ((\textbf{g}_t^k)^\text{CA})^2 $ 

    \STATE{}\ \ \ \ \ \  
    {$\textbf{m}^k_{t}$} $= \gamma_2$ {$\textbf{m}^k_{t-1}$} $+ (1-\gamma_2) (\textbf{g}_t^k)^\text{CA}$
   
    \STATE{}\ \   \textbf{End For}
    
    \STATE{}\ \  $\theta_t = \theta_{t-1} + \sum_k \Delta \theta^k_t$
    
    \STATE{} \textbf{End For}
\end{algorithmic}
\end{algorithm}
\begin{algorithm}[t]
\caption{{PCGrad(Adam) with AdaTask}}
\label{alg:xwithadatask_plus_pcgrad}
\small
\begin{algorithmic}[1]
    \STATE{\textbf{Require } Model parameters $\theta$}

    \STATE{} \textbf{For} step $t \in \{1,2,\ldots,T\}$ \ \textbf{do}
    
    \STATE{} \ \  \textbf{For} task $k \in \{1,2,\ldots,K\}$ \ \textbf{do}
    
     \STATE{}   \ \ \ \ \ \ $(\textbf{g}_t^k)^\text{PC} \gets \textbf{g}_t^k$ \ $\forall k$
     
    \STATE{} \ \ \ \ \ \ \textbf{For} task $j \sim \{1,2,\ldots,K\} \setminus k $ \ \text{in random order} \textbf{do}
    
    \STATE{} \ \ \ \ \ \ \ \ \ \ \textbf{If} $(\textbf{g}_t^k)^\text{PC} \cdot \textbf{g}_t^j < 0$ \  \textbf{then}
    
    \STATE{} \ \ \ \ \ \ \ \ \ \ \ \ \ \ \textit{// Subtract the projection of } $(\textbf{g}_t^k)^\text{PC} \textit{into}$ $\textbf{g}^j$
    
    \STATE{} \ \ \ \ \ \ \ \ \ \ \ \ \ \ \text{Set} $(\textbf{g}_t^k)^\text{PC} = (\textbf{g}_t^k)^\text{PC} - \frac{(\textbf{g}_t^k)^\text{PC} \odot \textbf{g}_t^j}{\|\textbf{g}_t^j\|^2} \textbf{g}_t^j$
    
    \STATE{} \ \ \ \ \ \ \ \ \ \ \textbf{End If} 
    
    \STATE{} \ \ \ \ \ \  \textbf{End For} 
    
    \STATE{} \ \ \ \ \ \  $\Delta \theta^k_{t}$   {$\gets - \frac{\eta}{\sqrt{\hat{\textbf{G}}^k_{t}} + \epsilon} \hat{\textbf{m}}^k_{t} $}  
        
    \STATE{}\ \ \ \ \ \   \text{where } 
    
    \STATE{}\ \ \ \ \ \  \ \   {$\hat{\textbf{G}}^k_{t}$} = {${\textbf{G}}^k_{t}$} $/ (1-\gamma_1^t)$, \;\; 
    {$\hat{\textbf{m}}^k_{t}$} = {${\textbf{m}}^k_{t}$} $/ (1-\gamma_2^t)$ 

    \STATE{}\ \ \ \ \ \ \ \  
    {$\textbf{G}^k_{t}$} $= \gamma_1$ {$\textbf{G}^k_{t-1}$} $+ (1-\gamma_1)
    ((\textbf{g}_t^k)^\text{PC})^2 $ 

    \STATE{}\ \ \ \ \ \ \ \  
    {$\textbf{m}^k_{t}$} $= \gamma_2$ {$\textbf{m}^k_{t-1}$} $+ (1-\gamma_2) (\textbf{g}_t^k)^\text{PC}$
    
    \STATE{}\ \ \ \  \textbf{End For}
    
    \STATE{}\ \  $\theta_t = \theta_{t-1} + \sum_k \Delta \theta^k_t$
    
    \STATE{} \textbf{End For}
\end{algorithmic}
\end{algorithm}

\section{Appendix D: Task Dominance Analysis}
\subsection{Task Dominance on the CityScapes}
In the main paper, we analyze the task dominance of multiple MTL methods on the synthetic dataset (i.e., Fig.~\ref{fig:EW_GN_AdaTask_toy_ratio}). Here we present the task dominance analysis on real-world datasets. We choose EqualWeight, GradNorm and AdaTask in the following analysis, and present the results in Fig.~\ref{fig:cityscapes_au_dominance}.

We observe that $99\%$ parameters in EqualWeight are dominated by task $B$; that is, $\text{rAU}(i,T,B)$ of almost all parameters belongs to (80\%, 100\%], which is the red area in Fig.~\ref{fig:cityscapes_au_dominance}(a). In addition, the GradNorm (Fig.~\ref{fig:cityscapes_au_dominance}(b))  has 95\% of the parameters dominated by task $B$. However, only 7\% of shared parameters in AdaTask(Fig.~\ref{fig:cityscapes_au_dominance}(c)) are dominated, i.e., 93\% of $\text{rAU}(i,T,B)$ belongs to (40\%, 60\%]. These results verify that most parameters of EqualWeight and GradNorm are dominated by one task in real-world datasets, while AdaTask solves it decently.

\subsection{Task Dominance on the TikTok}
As shown in Fig.~\ref{fig:tiktok_au_dominance_mlp} and Fig.~\ref{fig:tiktok_au_dominance_emb}, we analyzed the dominance of EqualWeight, GradNorm, and AdaTask in the sharing layer (divided into Embedding layer and MLP layer) on the TikTok dataset in the recommender system field. At the MLP layer (Fig.~\ref{fig:tiktok_au_dominance_mlp}), we observe a similar situation with the synthetic dataset and the CityScapes dataset, that is, AdaTask balances the shared parameters well. However, in the embedding layer (Fig.~\ref{fig:tiktok_au_dominance_emb}), we observed that AdaTask does not fully balance the parameters. We leave it for future work.

\subsection{Layer-wise Task Dominance on the Synthetic Dataset}

\begin{figure*}[h]
    \centering
    \setlength{\abovecaptionskip}{0.cm}
    \includegraphics[width=1\linewidth]{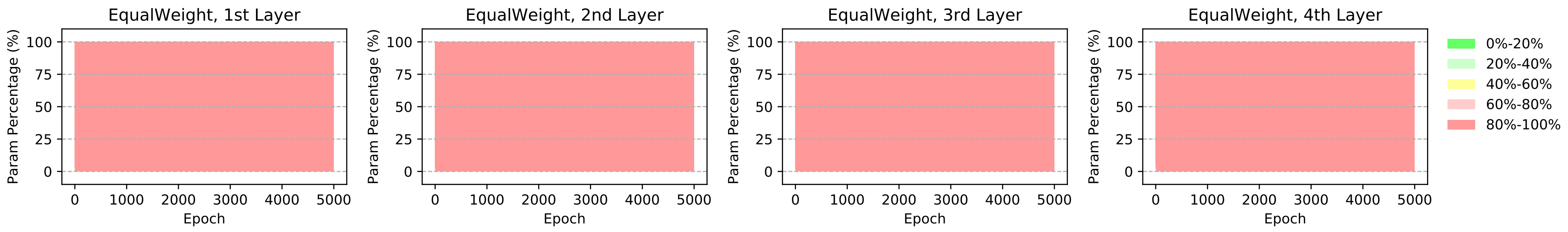}
    \includegraphics[width=1\linewidth]{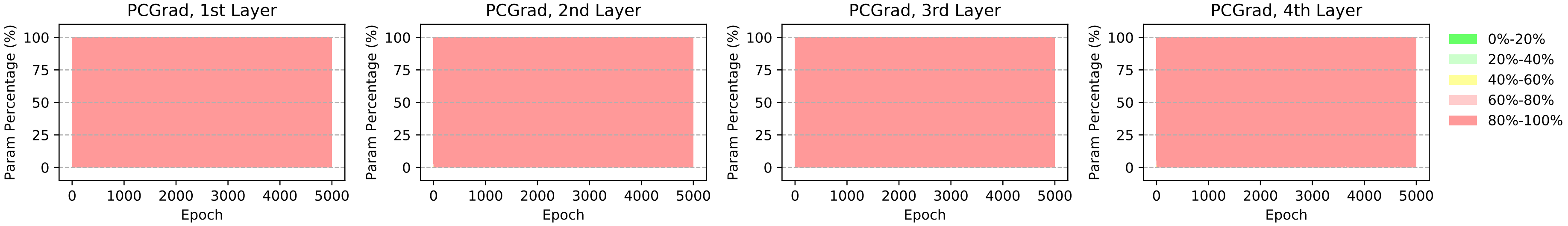}
    \includegraphics[width=1\linewidth]{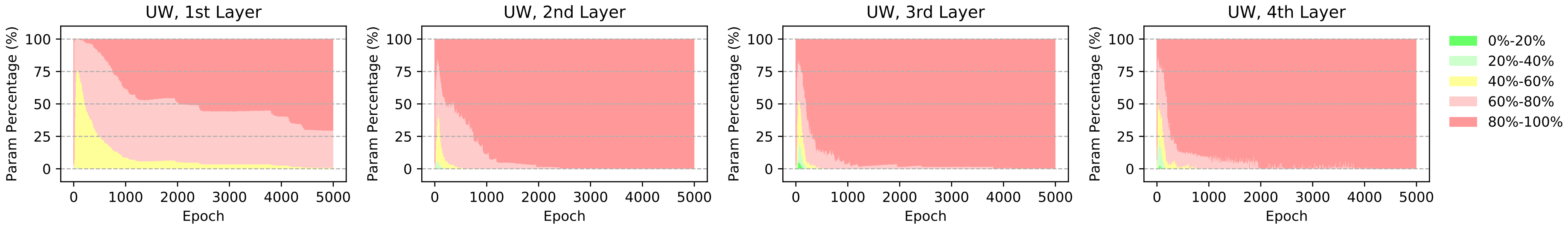}
    \includegraphics[width=1\linewidth]{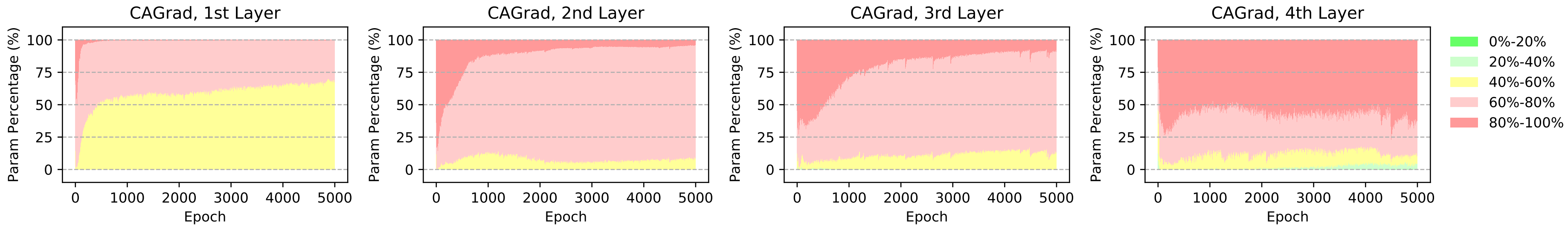}
    \includegraphics[width=1\linewidth]{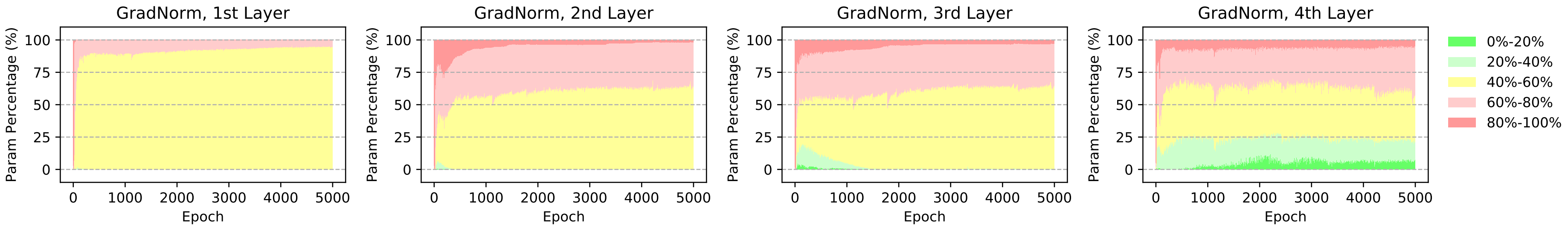}
    \includegraphics[width=1\linewidth]{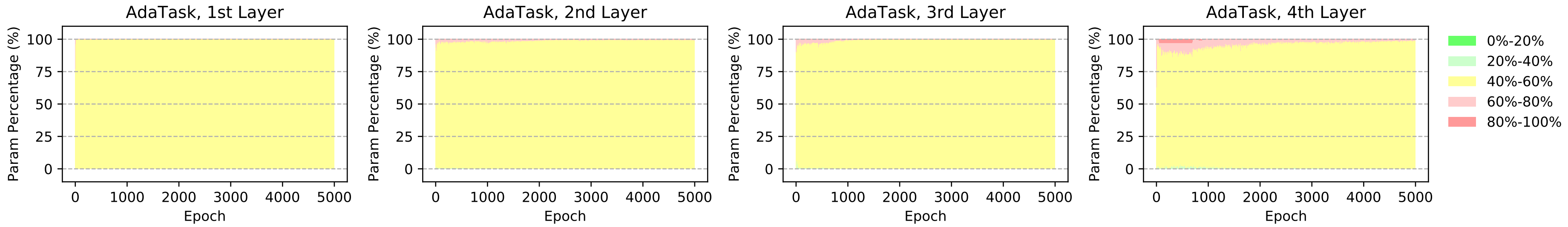}
    \caption{$\text{rAU}(i,T,B)$ of shared parameters in each layer on the synthetic dataset for six MTL methods:     EqualWeight, PCGrad, UW, CAGrad, GradNorm, and our AdaTask.}
    \label{fig:EW_GN_AdaTask_toy_layer_ratio}
\end{figure*}

In this section, we present evaluation results of AdaTask. As shown in Fig.~\ref{fig:EW_GN_AdaTask_toy_layer_ratio}, PCGrad is similar to EqualWeight in the sense that they are completely dominated by task $B$ at each layer. UW, CAGrad, and GradNorm are balanced better, especially at the bottom 
layer (1st layer), while they are still dominated at the top layers. In contrast, AdaTask achieves a good balance across all layers.

\subsection{Dominance of Learning Rates on the Synthetic Dataset}
In this section we demonstrate the dominance of learning rates in MTL. Similar to the parameter cumulative gradient-dominated analysis in the main text, we analyze the learning rate-dominated problem on synthetic dataset (Eq.~\ref{eq:synthetic}) based on the SharedBottom model (containing four shared fully-connected layers).
We randomly select ten parameters from each shared layer and calculate the learning rates calculated using their own accumulative gradients (i.e., \texttt{LR-A} and \texttt{LR-B}) and the learning rates calculated using the overall gradients accumulated across both tasks (i.e., \texttt{LR-Whole}).

Fig.~\ref{fig:sythetic_learning_rate_dominance}(a) presents the learning rate of parameters of each layer in the training process. We observe that: \texttt{LR-B} is much closer to \texttt{LR-Whole} than \texttt{LR-A} because the overall accumulative gradients are dominated by task $B$. Task $A$ should have a larger learning rate based on its own accumulative gradients. 
As shown in Fig.~\ref{fig:sythetic_learning_rate_dominance}(b), our proposed AdaTask adjusts the learning rate by using its own accumulative gradients, therefore there is no dominance on learning rates at all.

\section{Appendix E: Training Time}
In this section, we evaluate the training time cost by running each method for one epoch on the CityScapes dataset. As shown in Table~\ref{tab:timecost}, it can be observed that the training time of AdaTask is less than all other MTL methods except EqualWeight. This is because, compared to other MTL approaches, AdaTask only needs to separate the accumulative gradients, which introduces little additional computation cost.
\begin{table}[h]
\centering
\begin{tabular}{l|c c c}
\hline
Method    & EqualWeight & PCGrad   & GradNorm   \\ 
Time cost & 249.20s    & 343.01s & 346.51s  \\ \hdashline
Method    & CAGrad  & MGDA  & AdaTask (ours) \\ 
Time cost & 353.64s & 442.62s & 332.14s \\ \hline
\end{tabular}
\caption{Each method performs the time cost comparison of a single Epoch on the CityScapes dataset.}
\label{tab:timecost}
\end{table}

\section{Appendix F: Adaptive Learning Rate Methods (RMSProp and Adam) v.s. SGD Method}
In this section, we compare the performance of the adaptive learning rate methods and the global learning rate method on the CityScapes dataset.
Like CAGrad~\cite{cagrad-liu2021conflict}, we use average per-task performance improvement to measure the overall performance improvement of RMSProp and Adam over SGD.
As shown in Table~\ref{tab:cityscapes_sgdoptimizer}, we find that adaptive learning rate approaches (RMSProp and Adam) perform better than the SGD. Specifically, in the CityScapes dataset, we observe a $28.33\%$ increase in RMSProp relative to SGD and a $27.86\%$ increase in Adam relative to SGD.  
However, as pointed out in this paper, these traditional adaptive learning rate optimizers accumulate gradients across tasks in multi-task learning. Therefore they still suffer from accumulative gradient dominance.

\begin{table}[t]
\small 
\centering
    {%
    \begin{tabular}{lcrrrrc}
     \hline
     & \multicolumn{2}{c}{{Task A}} &\multicolumn{2}{c}{{Task B}} & {Rel Avg}  \\
    \cmidrule(lr){2-3}\cmidrule(lr){4-5}
    {Optimizer} & Abs$\downarrow$ & Rel$\downarrow$ & mIoU$\uparrow$  & Pix$\uparrow$  & $\Delta p$$\uparrow$  \\
      \hline 
     SGD      &  0.0242 &  115.35 & 66.51 & 90.89  &  0.00\%\\  \hdashline
     RMSProp  &  0.0152 &  44.63  & 74.86 & 93.43  & +28.33\% \\    
     Adam     &  0.0152 &  47.00  & 75.01 & 93.40  & +27.86\% \\   \hline
    \end{tabular}
    }
\caption{Performance comparison of adaptive learning rate approaches and SGD on CityScapes dataset.
\label{tab:cityscapes_sgdoptimizer}
\vspace{-15pt}
}
\end{table}

\section{Appendix G: Algorithm Pseudocode}
PCGrad~\cite{gradientsurgery} and CAGrad~\cite{cagrad-liu2021conflict} are two gradient direction modification methods, and they are orthogonal to  our proposed AdaTask and therefore they can be combined with AdaTask. We present the pseudocode of PCGrad+AdaTask and CAGrad+AdaTask in Alg.~\ref{alg:xwithadatask_plus_pcgrad} and Alg.~\ref{alg:xwithadatask_plus_cagrad}, respectively. Their performances are reported in Table~\ref{tab:cityscapes_mtan}.

\end{document}